\documentclass{article}
\usepackage{iclr2027_conference,times}
\usepackage[T1]{fontenc}
\usepackage{amsmath,amsfonts,bm}

\def\eqref#1{equation~\ref{#1}}

\def\1{\bm{1}}

\DeclareMathAlphabet{\mathsfit}{\encodingdefault}{\sfdefault}{m}{sl}
\SetMathAlphabet{\mathsfit}{bold}{\encodingdefault}{\sfdefault}{bx}{n}

\usepackage{hyperref}
\usepackage{url}
\usepackage{longtable}
\usepackage{wrapfig}
\usepackage{booktabs}
\usepackage{graphicx}
\usepackage{multirow}
\usepackage[table]{xcolor}

\title{Mimir: A Sub-2B Hierarchical Reasoning Model Trained on Permissible Data Punches Above Its Weight Class}
\author{Peter Schneider-Kamp$^{1,2,\ast}$ \quad Jacob Nielsen$^{1,2}$ \quad Gianluca Barmina$^{1}$ \quad Danial Namazifard$^{1}$\\
\textbf{Annemette Brok Pirchert}$^{1}$ \quad \textbf{Kenneth Enevoldsen}$^{3}$ \quad \textbf{Lukas Galke Poech}$^{1}$ \\
$^{1}$University of Southern Denmark \quad $^{2}$Ordbogen A/S \quad $^{3}$Aarhus University
}

\iclrfinalcopy
\begin{document}
\maketitle
\begin{abstract}
Training capable language models typically relies on web-scale pretraining followed by multi-stage post-training, creating substantial computational and data-governance barriers for settings in which training data must satisfy explicit permissibility constraints. We introduce Mimir, a 1.78B-parameter language model based on the Hierarchical Reasoning Model Text (HRM-Text) architecture, a recurrent-depth transformer whose layers are split into a low-level (L) and a high-level (H) stack that are applied repeatedly in an interleaved schedule. Mimir is trained from random initialization directly on a curated mixture of instruction, reasoning, tool-use, multilingual, and synthetic data: 161 datasets and 70.5B tokens per epoch for Mimir v1\footnote{\url{https://huggingface.co/danish-foundation-models/DFM-Mimir}}, extended to 103B tokens per epoch for its continued-training successor Mimir v1.5\footnote{\url{https://huggingface.co/danish-foundation-models/DFM-Mimir-v1.5}}, with synthetic replacements for sources that do not satisfy permissibility criteria. Across 20 benchmarks in English, Danish (a low-resource language with $\sim$6M speakers), and Math \& Code, Mimir v1.5 attains the highest average score of all sub-2B models in every domain, outperforming both HRM-Text 1B and the looped transformers Ouro 1.4B and Ouro 2.6B, and Mimir matches or exceeds the performance of models of up to 9B parameters, such as Qwen 3.5 9B and Gemma 4 E4B, on the English and Danish averages. Two independent memorisation audits indicate low memorisation risk despite multi-epoch training. Inference-time ablations show that the number of L-stack applications can be changed with graceful degradation, whereas changing the number of H-stack applications breaks the model. Our results provide evidence that hierarchical recurrent language models can learn effectively from post-training-oriented, permissible data mixtures without conventional web-scale pretraining.
\end{abstract}
\section{Introduction}
\begin{figure}[htbp]
\centering
\includegraphics[width=0.99\textwidth]{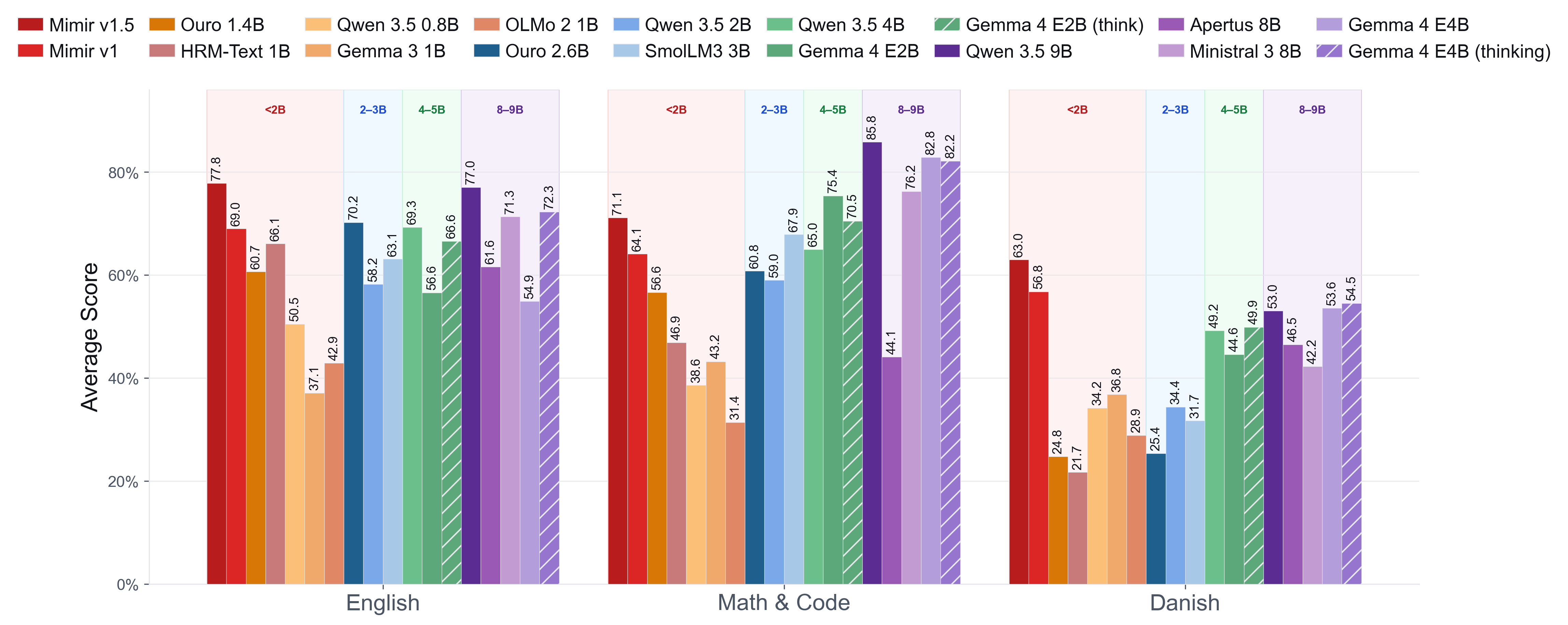}
\caption{Aggregate results across three categories English --  Math \& Code, and Danish -- comparing Mimir v1 and v1.5 against frontier models from 0.8B to 9B, such as HRM-Text 1B, Gemma 4 E4B, and Qwen 3.5 9B, displaying highly competitive performance across 20 benchmarks against both frontier models and leading European private and public efforts.}
\label{fig:model-comparison-benchmark}
\end{figure}

Current developments in LLM training are largely driven by a monolithic recipe consisting of massive, multi-stage pipelines trained on exorbitant volumes of data~\citep{wang2026hrm}. This recipe requires vast compute for pre-training and creates a prohibitive entry barrier for the broader community of researchers and practitioners. This barrier is especially high for efforts, including several in Europe, that commit to
regulation, copyright law, and the use of exclusively permissible and ideally openly licensed data: the pool of such high-quality data is limited for most languages, and even where data exists, pre-, mid-, and post-training demand massive compute. Training capable, instruction-ready language models from scratch under these constraints therefore remains a key challenge.
Tackling this challenge, we present \textbf{Mimir}, which combines the HRM-Text architecture~\citep{wang2026hrm} with scalable instruction data pipelines for training. The recurrent depth architecture, along with a PrefixLM attention mask enable focusing on post-training data from the beginning of training. As a result, Mimir is a 1.78B hierarchical reasoning model, trained from scratch, that leads the Pareto frontier on both English and Danish, outperforms similar recurrent depth models, and competes with larger models of up to 9B parameters (Figure~\ref{fig:model-comparison-benchmark}).
In more detail, Mimir v1 is trained on a curated mixture of 161 English and Danish datasets comprising approximately 70.5B tokens per epoch, a five-fold increase over the training data of HRM-Text 1B.
Furthermore, as certain datasets used in \citep{wang2026hrm} do not align with our permissibility standards, we demonstrate the efficacy of synthetically generating ``transplant'' datasets, replacing non-permissible data with synthetically generated permissible variants. With module-specific learning rate scaling as well as expansion to 103B tokens per epoch, iterative auditing, and refinement of the datasets, we continually train the model to attain Mimir v1.5. Mimir v1.5 is best-in-class amongst recurrent depth approaches and competes with larger frontier models of sizes up to 9B.
The core of our contribution is a set of synthetic instruction-data pipelines with automated auditing, from which we assemble the Mimir Training Corpus (70.5B tokens per epoch for v1 and 103B for its continued-training successor v1.5). We combine HRM-Text with the Gemma 4 tokenizer and chat template, and gradually extend backpropagation through the recurrence over the course of training.
Our results show that our synthetic data pipeline leads to comparable or superior performance without compromising data rights, further underscoring the relevance of HRMs for low-resource language modeling, empowering communities of practitioners and researchers with small, capable and fully permissible models, with both low training and inference budgets.
We evaluate Mimir on 20 standard benchmarks in English, Math \& Code, and Danish. Two independent memorisation audits find that the fraction of memorised responses is negligible, indicating low copyright risk under the tested scenarios. Finally, inference-time ablations of the recurrence show that the number of low-level (L) module applications is elastic, whereas the number of high-level (H) applications is not.
In sum, our contributions are as follows:
\begin{itemize}
    \item  We introduce \textbf{Mimir} (\S\ref{sec:model}), an HRM-based language model trained from random initialization directly on a permissible mixture of instruction, reasoning, multilingual, tool-use, and synthetic data (161 datasets, 70.5B sampled tokens per epoch for Mimir v1), together with Mimir v1.5, which has been continually trained on 103B tokens per epoch.
    \item We introduce a synthetic training data generation pipeline (\S\ref{sec:data}) through which we create the Mimir Training Corpus comprising 161 datasets. The pipeline includes synthesis of grammar correction, denoising, and paragraph reordering tasks, and Mimir only receives gradient signal from the completions.
    \item We provide an extensive evaluation across English, Math \& Code, and Danish benchmarks, showing Mimir represents the new state of the art among sub-2B models, and competes with larger frontier models such as Gemma 4 E4B and Qwen 3.5 9B (\S\ref{sec:evals}).
    \item We further report on
    (a) Mimir's memorisation behavior, including extraction audits of data sources with potential copyright risk (\S\ref{sec:audits}); and (b) inference-time ablations that vary both the total recurrent depth and its split between the H- and L-modules (\S\ref{sec:ablations}).
\end{itemize}
\section{Data}\label{sec:data}
\subsection{Synthetic Instruction Data Pipeline}\label{sub:synthetic}
Satisfying our permissibility criteria rules out important parts of the data used to train HRM-Text 1B~\citep{wang2026hrm}, and permissible Danish instruction data is scarce. We therefore build six complementary pipelines that turn permissible sources
into instruction-tuning data. All pipelines share a generate-and-audit design: Gemma 4 31B~\citep{gemmateam2026gemma4technicalreport} generates candidate examples and, in a separate call, audits each candidate against pipeline-specific criteria described below.
Only accepted examples enter the Mimir training corpus.
Further details are in Appendix~\ref{app:mtc}.
\paragraph{Turning text corpora into instruction data.}
Openly licensed text corpora contain rich language but are not in the instruction format Mimir is trained on. We convert texts into supervised tasks, for which the answers originate from the source text itself, i.e., the generator must use text that exists verbatim in the corpus and the auditor checks this.
Specifically, we construct four task families: corruption repair (denoising), T5-style span filling, causal-style prefix continuation, and paragraph reordering. We apply these transformations to documents from the Common Pile~\citep{kandpal2025commonpilev018tb} for English and from Danish Dynaword~\citep{6c685239843b47bfb7f47a78aae67929} for Danish. These instruction data contribute 3.02B and 0.99B tokens per epoch, respectively. The rationale for this design is that it grounds a substantial part of the training signal in human-written text
, while the answers remain verifiable against the source.
These tasks also suit the PrefixLM attention mask: the corrupted, masked, truncated, or reshuffled text sits in the bidirectionally attended prefix, and the loss is computed only on the original text in the response.
\paragraph{Repairing instead of filtering instruction datasets.}
Existing instruction datasets often contain flawed answers, including factually incorrect, irrelevant or improperly formatted ones.
We audit each conversation and either keep it, regenerate the flawed answer, or discard it if it cannot be repaired; each regenerated answer is re-audited against the original conversation, so that repairs do not drift from the intent of the prompt.
For the English OpenHermes~\citep{OpenHermes2.5} data, this keeps 460,695 of 1,001,551 conversations unchanged (46.0\%), replaces 457,400 with accepted repairs (45.7\%), and discards 83,456 (8.3\%).
We record row-level evidence for every decision.
\paragraph{Structure-preserving translation into Danish.}
To extend high-quality English instruction data to Danish, we machine-translate conversations with the generator model while preserving their structure: roles, code, formulas, labels, and structured outputs remain intact, and only the natural-language content is translated. The rationale is that naive translation often corrupts the elements that make instruction data useful, such as code blocks or answer labels that must match a fixed format. We audit each translation, and whenever the English source is repaired (see above), we retranslate the repaired source and replace the corresponding Danish row, which keeps the two languages aligned. This pipeline yields the Danish OpenHermes data (0.92B tokens per epoch).
\paragraph{Deriving task-specific supervision from domain documents.}
Agreement-supplied Danish sources, i.e., DBC and Lex.dk, are of high quality but may not be shared publicly. We therefore do not train on the documents directly, but derive instructions from each document's natural structure.
For example, book descriptions and reviews yield summarization and characterization tasks, encyclopedia articles yield question answering, and paired documents yield transformation tasks.
Because the model only sees derived tasks, this conversion also limits the surface for verbatim memorisation of the source documents (as we tested in our memorization audits in Sec.~\ref{sec:audits}).
\paragraph{Targeting observed data needs.}
Small models with limited capacity benefit more from data aimed at specific needs than from additional generic instructions~\citep{wang2025slmsurvey}. In order to target such needs, our synthetic data generation approach is guided by failures observed on intermediate checkpoints.
We identify task families and prompt variations relevant for addressing such failures, then generate and audit bilingual examples for each family. This results in six new families: native tool calling, Danish summarization with rewrite controls, multi-turn Danish--English chat, code debugging, strict math answer formats, and constrained format following.
\paragraph{Converting heterogeneous agent traces into native tool use.}
Agentic datasets encode tool calls in incompatible formats, often wrapped in legacy conventions that conflict with newer chat templates. We convert these traces into native tool-use supervision by recovering the declared tool schemas, the structured calls, the matching tool results, and the final answers. We remove incompatible legacy wrappers and repair malformed trajectories, for example calls without a matching result.
We apply this conversion to Dolci~\citep{olmo2025olmo3}, Glaive\footnote{Hugging Face ID: \texttt{glaiveai/glaive-function-calling-v2}}, xLAM~\citep{liu2024apigen}, ToolACE~\citep{liu2025toolace}, and Nemotron~\citep{nvidia2025nemotron3nano} for Mimir v1, and additionally to DeepDive~\citep{lu2025deepdive} for Mimir v1.5.
\subsection{Mimir Training Corpus v1}
The final training corpus for Mimir v1 covers varying objectives, ranging from English and Danish instruction and knowledge to mathematics and agentic-style post-training data. This mixture, which we denote the Mimir Training Corpus v1, draws on 161 datasets with all but 2 Danish ones being freely available on the Hugging Face Hub.  The corpus amounts to \textbf{70{,}479{,}308{,}606 tokens} per epoch.
Appendix~\ref{app:list-of-datasets} lists all datasets and their sources.
\paragraph{Language Distribution.}
Table~\ref{tab:language} shows the distribution of tokens by language.
The corpus is predominantly English (68.6\%), with Danish contributing 24.7\%, bilingual Danish--English data a further 6.5\%, and translation data between Danish and other languages 0.2\%.
\paragraph{Category Distribution.}
We classify each dataset into one of eight functional categories based on its content and intended use. Table~\ref{tab:category} shows that 83\% of tokens come from outside the HRM-Text collection, and that the three largest of these categories focus on generative tasks: \textbf{English instruction} (\(13.58\text{B}\), \(19.3\%\)), \textbf{Math \& reasoning} (\(10.40\text{B}\), \(14.8\%\)), \textbf{Danish instruction \& knowledge} (\(15.56\text{B}\), \(22.1\%\)).
These three categories account for 39.54B tokens (56.1\% of the corpus), nearly all relying on answer or free-form generation rather than multiple-choice selection.
Appendix~\ref{app:mtc} provides further details on the Mimir Training Corpus v1.
\subsection{Mimir Training Corpus v1.5}
The Mimir Training Corpus v1.5 extends v1 to \textbf{103{,}214{,}604{,}702} tokens per epoch, 46\% more than v1. It replaces instruction datasets flagged by an audit of 100 random samples per dataset with repaired versions.
All but the same two Danish datasets are available from the Hugging Face hub.
\begin{table}
\small
\centering
\caption{Token share and dataset count by functional category.}\label{tab:category}
\tiny
\begin{tabular}{lrrrr}
\toprule
\textbf{Category} & \textbf{Tokens/epoch} & \textbf{Share} & \textbf{Datasets} & \textbf{Avg. tokens/dataset} \\
\midrule
Danish instruction \& knowledge & 15.56B & 22.07\% & 30 & 518.6M \\
English instruction            & 13.58B & 19.26\% & 16 & 848.5M \\
Sapient mixed (Flan/Platypus)  & 12.00B & 17.02\% & 71 & 169.0M \\
Math \& reasoning              & 10.40B & 14.76\% & 10 & 1{,}040.2M \\
Mimir synthetic                 & 7.05B  & 10.00\% & 16 & 440.4M \\
Agentic \& tool use            & 6.66B  & 9.46\%  & 8  & 832.5M \\
Machine translation            & 3.50B  & 4.96\%  & 5  & 699.0M \\
Science \& summarization       & 1.74B  & 2.47\%  & 5  & 348.5M \\
\midrule
\textbf{Total}                  & \textbf{70.48B} & \textbf{100\%} & \textbf{161} & \\
\bottomrule
\end{tabular}
\end{table}
\subsection{Comparison to HRM-Text Data}
Notably, the training data for HRM-Text 1B\footnote{Hugging Face ID: \texttt{sapientinc/HRM-Text-data-io-cleaned-20260515}} is dominated by multiple-choice classification tasks: Flan, Platypus, and tasksource sub-collections.
This is a natural consequence of their source benchmarks: Flan NIV2, Platypus (Reclor, SciBench), and task source (PragmEval, ReClor), which are structured around categorical selection.
For Mimir, we shift from multiple-choice toward free-form generation, presenting a substantially harder task for the model to achieve exact match accuracy.
This change in data composition better prepares Mimir for open-ended conversation and free-form generation, favoring direct-answer tasks over multiple-choice option selection.
\section{Model}\label{sec:model}
\paragraph{Architecture.}
Mimir follows the Hierarchical Reasoning Model Text (HRM-Text) architecture~\citep{wang2026hrm,wang2025hierarchicalreasoningmodel}, an instance of the broader class of Looped Transformer models~\citep{giannou2023looped,yang2024looped,ICLR2025_2676109d}, whereby the same transformer modules are applied repeatedly to iteratively refine the residual stream representation. The architecture is designed in a way that is reminiscent of Kahneman's ideas of thinking, fast and slow~\citep{kahneman2011thinking}. It consists of two Transformer stacks, both with a PrefixLM attention mask: one L-module for fast, low-level and one H-module for slower, high-level reasoning. To operationalize this idea, the L-module is applied more frequently, in consecutive runs before each H-module application (e.g., three L applications for each of two H applications, denoted as \texttt{LLLHLLLH}). The PrefixLM attention mask enables bidirectional attention within the instruction or prompt prefix, while preserving the causal structure for the response to facilitate generation.
The states are initialized as $z_H^{0}=x$ (the input embedding) and $z_L^{0,0}=z_{L,\mathrm{init}}$ (a learned state). For each high-level cycle $i$, the L-module is applied $N_L$ times, $z_L^{i,j+1} = f_L(z_L^{i,j} + z_H^{i})$ for $j=0,\dots,N_L-1$, after which the H-module updates $z_H^{i+1} = f_H(z_H^{i} + z_L^{i,N_L})$ and the next cycle continues from $z_L^{i+1,0}=z_L^{i,N_L}$.
The L module takes as input a combination of the previous state of the residual stream and the last-updated H representation. The input embeddings initialize the H-level state, effectively connecting the L module to the input.
Mimir employs a hidden size of 1{,}536. The model has 12 attention heads per layer and a feed-forward expansion factor of 4. We employ a Rotary Position Embedding (RoPE)~\citep{su2024roformer} with $\theta = 10{,}000$ and pre-normalization within each block, applying the RMSNorm before both the self-attention and the feed-forward sublayers with $\epsilon = 10^{-6}$. An additional RMSNorm is applied after the final Transformer block before the output head. We provide the full model configuration in Appendix~\ref{app:hparams:model}. Mimir has 1.38B non-embedding parameters and 0.40B embedding parameters.
For Mimir v1, backpropagation through the recurrence is truncated to the last five module applications (\texttt{HLLLH}); the truncation window grows linearly from two to five applications over the first 20\% of optimizer steps.
We use the Gemma 4 tokenizer~\citep{gemmateam2026gemma4technicalreport}, with a 262{,}144-token vocabulary, and its chat template, which provide efficient multilingual tokenization and native conventions for multi-turn chat and tool calling.
\paragraph{Training Mimir v1.}
Mimir v1 was trained with Fully Sharded Data Parallelism using bfloat16 as dType for computation and fp32 as gathering precision. We employ the AdamW optimizer~\citep{loshchilov2017decoupled} with a peak learning rate of  $3 \times 10^{-4}$, 2,000-step linear warm-up, and a constant schedule thereafter. We use a global batch size of 262{,}144 tokens with a gradient accumulation of 2 on 8 accelerators for a per-accelerator batch size of 16,384, fitting 4 contexts of length 4,096 each. Appendix~\ref{app:hparams:train} provides an overview of training hyperparameters.
\noindent
We trained Mimir v1 for $1.65$M steps on 8 NVIDIA B200 GPUs with 180 GB HBMe3 in just under 3 weeks with an average step time of just under 1.1 seconds, i.e., using approx.\ 8,000 B200 hours.
\paragraph{Continued Training: Mimir v1.5.}
We continually trained Mimir for $1.2M$ steps on the Mimir Training Corpus v1.5.
For Mimir v1.5, we gradually expand the backpropagation steps from 5 steps for Mimir v1 to full backpropagation, i.e., all 8 steps of the \texttt{LLLHLLLH} recurrence.
Because the $H$- and $L$-modules reuse their weights across iterations, a single parameter update to a module changes the model's output through every one of its applications.
We therefore divide the base learning rate $\eta$ by the number of applications of each module that receive gradients, i.e., $\eta_M = \eta / k_M$ for $M \in \{H, L\}$, where $k_M$ is determined by the current backpropagation window. For example, with \texttt{LLLHLLLH} recurrence and backpropagation truncated to the last five steps (i.e., \texttt{HLLLH}), the $H$-module receives two gradient contributions and the $L$-module three, which yields $\eta_H = \eta/2$ and $\eta_L = \eta/3$.  Specifically, when the truncation window grows from two to five updates during the first 20\% of training, the dividers follow the window: at two steps (\texttt{LH}), both modules are trained with $\eta$, whereas at four steps (\texttt{LLLH}), $\eta_L = \eta/3$ and $\eta_H = \eta$. The peak learning rate of $3 \times 10^{-4}$ refers to $\eta$ before division. We found that this scaling was critical to stabilize training when increasing the backpropagation to all 8 steps of the recurrence.
We close Mimir v1.5 training with a cooldown epoch scaling down the base learning rate to $3 \times 10^{-5}$.
\begin{table}[htbp]
\centering
\caption{English benchmark results. Best scores in \textbf{bold}.}
\label{tab:english_summary}
\vspace{6pt}
\scriptsize
\setlength{\tabcolsep}{4pt}
\begin{tabular}{llccccccc|r}
\toprule
\textbf{Band} & \textbf{Model} & \textbf{BoolQ} & \textbf{Winogrande} & \textbf{Hellaswag} & \textbf{MMLU} & \textbf{ARC-C} & \textbf{DROP} &  \textbf{GovRep.} & \textbf{\emph{Avg.}}\\
 & & (Acc) & (Acc) & (Acc) & (Acc) & (Acc) & (F1) & (R1) \\
\midrule
\rowcolor{gray!25}
\multirow{7}{*}{$<$2B} & Mimir v1.5                & \textbf{91.6} & \textbf{81.8} & 83.9 & 67.4 & 90.4 & \textbf{88.8} & \textbf{40.9} & \textbf{77.8} \\
\rowcolor{gray!25}
& Mimir v1                  & 87.8 & 73.5 & 67.3 & 57.5 & 81.6 & 83.1 & 32.0 & 69.0 \\
 & Ouro 1.4B                 & 83.9 & 56.0 & 59.6 & 67.1 & 87.0 & 38.9 & 32.4 & 60.7 \\
 & HRM-Text 1B               & 87.5 & 70.4 & 60.4 & 58.7 & 82.2 & 78.1 & 25.4 & 66.1 \\
 & Qwen 3.5 0.8B             & 69.8 & 48.9 & 37.0 & 51.5 & 68.4 & 45.2 & 32.5 & 50.5 \\
 & Gemma 3 1B                & 62.4 & 49.1 & 30.6 & 37.5 & 43.5 & 7.0 & 29.5 & 37.1 \\
 & OLMo 2 1B                 & 67.2 & 51.0 & 42.4 & 41.6 & 48.1 & 12.4 & 37.7 & 42.9 \\
\midrule
\multirow{3}{*}{2--3B} & Qwen 3.5 2B          & 80.8 & 53.4 & 64.6 & 62.8 & 82.7 & 31.3 & 31.5 & 58.2 \\
 & SmolLM3 3B                & 84.3 & 60.3 & 65.1 & 60.2 & 79.5 & 54.0 & 38.1 & 63.1 \\
 & Ouro 2.6B                 & 89.0 & 71.7 & 78.4 & 74.5 & 93.0 & 52.0 & 33.1 & 70.2 \\
\midrule
\multirow{3}{*}{4--5B} & Qwen 3.5 4B               & 87.0 & 70.0 & 83.2 & 75.8 & 92.9 & 48.0 & 27.9 & 69.3 \\
 & Gemma 4 E2B               & 64.1 & 56.7 & 55.6 & 59.3 & 69.8 & 57.3 & 33.6 & 56.6 \\
 & Gemma 4 E2B (thinking)       & 83.4 & 63.0 & 55.8 & 72.0 & 86.8 & 70.8 & 34.7 & 66.6 \\
\midrule
\multirow{5}{*}{8--9B} & Qwen 3.5 9B               & 89.3 & 75.1 & \textbf{88.6} & \textbf{79.5} & \textbf{94.1} & 82.1 & 30.7 & 77.1 \\
 & Apertus 8B                & 79.7 & 56.0 & 66.9 & 62.1 & 79.3 & 52.5 & 35.0 & 61.6 \\
 & Ministral 3 8B            & 88.2 & 67.2 & 67.6 & 73.9 & 90.1 & 81.0 & 31.3 & 71.3 \\
 & Gemma 4 E4B               & 85.2 & 64.0 & 44.6 & 35.4 & 49.8 & 74.0 & 31.2 & 54.9 \\
 & Gemma 4 E4B (thinking)    & 88.4 & 76.1 & 69.8 & 68.6 & 88.8 & 79.9 & 34.2 & 72.3 \\
\bottomrule
\end{tabular}
\end{table}
\section{Evaluations}
\label{sec:evals}
\begin{wraptable}{r}{0.62\linewidth}
\vspace*{-3ex}
    \centering
    \caption{Math \& Code benchmark results. Best scores in \textbf{bold}.}\label{tab:mathcode}
    \vspace{6pt}
\scriptsize
\setlength{\tabcolsep}{4pt}
\begin{tabular}{llccc|r}
\toprule
     \textbf{Band} & \textbf{Model} & \textbf{GSM8K}  & \textbf{MATH}  & \textbf{HumanEval} & \textbf{\emph{Avg.}}\\
      & & (Acc) & (Acc) & (Acc) \\
     \midrule
\rowcolor{gray!25}
\multirow{7}{*}{$<$2B} & Mimir v1.5 & 91.7 & 48.4 & 73.2 & 71.1 \\
\rowcolor{gray!25}
& Mimir v1 & 89.9 & 45.8 & 56.7 & 64.1 \\
 & Ouro 1.4B & 63.5 & 51.6 & 54.9 & 56.6 \\
 & HRM-Text 1B & 84.8 & 56.0 & 0.0 & 46.9 \\
 & Qwen 3.5 0.8B & 49.1 & 36.1 & 30.5 & 38.6 \\
 & Gemma 3 1B & 49.7 & 37.2 & 42.7 & 43.2 \\
 & OLMo 2 1B & 59.4 & 18.8 & 15.9 & 31.4 \\
\midrule
\multirow{3}{*}{2--3B} & Qwen 3.5 2B & 73.7 & 55.7 & 47.6 & 59.0 \\
 & SmolLM3 3B & 80.0 & 62.2 & 61.6 & 67.9 \\
 & Ouro 2.6B & 74.0 & 59.0 & 49.4 & 60.8 \\
\midrule
\multirow{3}{*}{4--5B} & Qwen 3.5 4B & 60.5 & 56.5 & 78.0 & 65.0 \\
 & Gemma 4 E2B & 88.3 & 64.2 & 73.8 & 75.4 \\
 & Gemma 4 E2B (thinking) & 90.3 & 49.1 & 72.0 & 70.5 \\
\midrule
\multirow{5}{*}{8--9B} & Qwen 3.5 9B & \textbf{95.5} & 74.2 & \textbf{87.8} & \textbf{85.8} \\
 & Apertus 8B & 66.7 & 25.4 & 40.2 & 44.1 \\
 & Ministral 3 8B & 91.7 & 61.4 & 75.6 & 76.2 \\
 & Gemma 4 E4B & 92.0 & 71.8 & 84.8 & 82.9 \\
 & Gemma 4 E4B (thinking) & 93.8 & \textbf{75.3} & 77.4 & 82.2 \\
 \bottomrule
 \end{tabular}
\vspace*{-10ex}
\end{wraptable}
\paragraph{Setup.}
All benchmarks were evaluated under greedy decoding with shuffle seed~4242 on full test sets. Some English benchmarks use few-shot prompting, with the number of shots following \cite{wang2026hrm}'s evaluation methodology. All Danish tasks are 0-shot. Further details on the setup and evaluation datasets can be found in Appendix~\ref{app:evals}.
\paragraph{Reference models.}
We compare Mimir against 14 models from 0.8B to 9B parameters, including but not limited to Qwen 3.5 0.8B/2B/4B/9B~\citep{qwen3.5}, Gemma 4 E2B/E4B, and view HRM-Text 1B, Ouro 1.4B/2.6B~\citep{zhu2025scaling} as our main baselines.
\paragraph{Results.}
Tables \ref{tab:english_summary}, \ref{tab:mathcode}, and \ref{tab:danish_summary} show evaluation results across English, Math \& Code, and Danish benchmarks, respectively.
Mimir v1.5 attains the highest average of all models on English and Danish, and the highest average among models up to 3B on Math \& Code.
\begin{table}[h]
\centering
\caption{Danish benchmark results. Best scores in \textbf{bold}.}
\label{tab:danish_summary}
\vspace{6pt}
\scriptsize
\setlength{\tabcolsep}{3.25pt}
\begin{tabular}{llcccccccccc|r}
\toprule
\textbf{Band} & \textbf{Model} & \textbf{Angry} & \textbf{DaLA} & \textbf{GEC} & \textbf{PIQA} & \textbf{Daisy} & \textbf{WikiQA} & \textbf{WMT} & \textbf{N.News} & \textbf{IFEval} & \textbf{Hellaswag-} & \textbf{\emph{Avg.}} \\
& & \textbf{Tweets} &&&&&&&&& \textbf{DA}\\
 & & (Acc) & (F1) & (EM) & (Acc) & (EM) & (EM) & (chrF) & (chrF) & (Acc) & (Acc) \\
\midrule
\rowcolor{gray!25}
\multirow{7}{*}{$<$2B} & Mimir v1.5                & 63.3 & \textbf{96.8} & \textbf{92.5} & 73.2 & \textbf{12.3} & \textbf{66.9} & 54.6 & 29.4 & 66.4 & \textbf{74.4} & \textbf{63.0} \\
\rowcolor{gray!25}
& Mimir v1                  & 67.4 & 96.1 & 85.6 & 53.7 & 9.6 & 66.8 & 53.9 & 35.9 & 63.9 & 35.3 & 56.8 \\
 & Ouro 1.4B                 & 18.8 & 35.6 & 0.3 & \textbf{86.1} & 0.5 & 12.5 & 27.7 & 19.0 & 22.6 & 25.0 & 24.8 \\
 & HRM-Text 1B               & 42.4 & 26.7 & 0.5 & 13.0 & 0.0 & 34.9 & 25.4 & 26.8 & 18.5 & 28.8 & 21.7 \\
 & Qwen 3.5 0.8B             & 53.8 & 51.0 & 0.7 & 56.5 & 0.7 & 41.6 & 37.8 & 35.3 & 39.6 & 25.0 & 34.2 \\
 & Gemma 3 1B                & 54.4 & 41.0 & 3.3 & 72.2 & 1.4 & 42.6 & 45.1 & 35.6 & 47.2 & 24.8 & 36.8 \\
 & OLMo 2 1B                 & 33.6 & 48.7 & 0.2 & 75.0 & 0.0 & 8.4 & 30.0 & 33.8 & 32.5 & 26.7 & 28.9 \\
\midrule
\multirow{3}{*}{2--3B} & Qwen 3.5 2B               & 61.6 & 36.4 & 8.0 & 25.0 & 2.5 & 49.4 & 45.6 & 34.9 & 56.1 & 24.7 & 34.4 \\
 & SmolLM3 3B                & 63.2 & 33.5 & 3.3 & 51.9 & 2.2 & 0.3 & 37.3 & 36.0 & 49.8 & 40.1 & 31.7 \\
 & Ouro 2.6B                 & 17.8 & 33.3 & 0.8 & 75.9 & 1.4 & 24.8 & 30.0 & 19.1 & 26.8 & 24.2 & 25.4 \\
\midrule
\multirow{3}{*}{4--5B} & Qwen 3.5 4B               & 69.1 & 50.1 & 42.6 & 70.4 & 4.7 & 57.1 & 52.1 & \textbf{37.0} & 73.7 & 34.7 & 49.2 \\
 & Gemma 4 E2B               & 64.6 & 56.7 & 36.9 & 46.3 & 5.6 & 44.1 & 55.2 & 35.7 & 75.5 & 25.6 & 44.6 \\
 & Gemma 4 E2B (think)       & 67.7 & 66.8 & 23.4 & 63.9 & 5.1 & 59.3 & 56.0 & 36.3 & \textbf{81.2} & 39.0 & 49.9 \\
\midrule
\multirow{5}{*}{8--9B} & Qwen 3.5 9B               & 68.9 & 70.1 & 52.0 & 62.0 & 8.5 & 50.4 & 54.8 & 22.3 & 70.8 & 70.7 & 53.1 \\
 & Apertus 8B                & 63.5 & 51.5 & 37.6 & 69.4 & 10.8 & 59.4 & 55.8 & 21.3 & 56.0 & 40.0 & 46.5 \\
 & Ministral 3 8B            & 63.9 & 58.9 & 17.9 & 63.0 & 7.3 & 48.7 & 50.2 & 18.4 & 48.6 & 45.5 & 42.2 \\
 & Gemma 4 E4B               & 69.7 & 63.0 & 49.4 & 71.3 & 7.9 & 53.5 & 57.3 & 22.0 & 77.3 & 64.3 & 53.6 \\
 & Gemma 4 E4B (thinking)    & \textbf{69.9} & 72.6 & 39.7 & 69.4 & 7.6 & 60.6 & \textbf{57.5} & 21.4 & 80.4 & 66.2 & 54.5 \\
\bottomrule
\end{tabular}
\vspace{2pt}
\end{table}
On English, Mimir v1.5 scores best of all models on BoolQ, Winogrande, DROP, and GovReport, and second to Qwen 3.5 9B on HellaSwag; its average narrowly exceeds that of Qwen 3.5 9B (77.8 vs.\ 77.1). On Math \& Code, it scores best of all models up to 5B on GSM8K and best up to 3B on HumanEval, but trails HRM-Text 1B and Ouro 1.4B on MATH. On Danish, Mimir v1.5 scores best of all models on DaLA, GEC, Daisy, WikiQA, and HellaSwag-DA; Mimir v1 also exceeds every non-Mimir model on DaLA, GEC, and WikiQA, and both versions exceed every other model's Danish average. Relative to Mimir v1, v1.5 improves on every benchmark except Angry Tweets and N.News. Mimir v1.5 outperforms HRM-Text 1B on every benchmark except MATH, and Ouro 2.6B on every benchmark except MMLU, ARC-C, MATH, and PIQA-da.
\section{Memorisation Audits}\label{sec:audits}
Training data duplication and multi-epoch training, as conducted for Mimir v1 and v1.5, may generally lead to an increase in memorisation~\citep{kandpal2022deduplicating,lee-etal-2022-deduplicating,carlini2023quantifying,slack-al-moubayed-2025-early}.
Relatedly, regulatory frameworks may distinguish between lawful access, legitimate training, and copyright law. As legitimate training and copyright are two separate matters, studying memorisation behavior is of interest to assess the model's copyright violation risk.
Therefore, we run two memorisation audits on Mimir v1 independently from each other to assess both its propensity and its capability to memorize training data. We cover four data categories: synthetic instruction-tuning data derived from agreement-backed sources (A), instruction-tuning data from Hugging Face sources with uncertain opt-out status (B), instruction-tuning data from Hugging Face sources with high confidence of no applicable opt-out (C), and other low-risk synthetic and reasoning post-training datasets (D). All remaining datasets are either covered by open licenses, in the public domain, or synthetic datasets derived from such openly licensed or public domain datasets.
The first audit with targeted prefix attacks yielded predominantly short matches, verbatim spans of 50 tokens or more were found in only 0.00022\%--0.015\% of training documents across all categories.
Across all model-input evaluations in the second audit, only 0.000045\% corresponded to coherent prose and 0.00000073\% to expressive prose, with no critical copyright findings, indicating low copyright-related memorisation risk under the tested scenarios. Details are in Appendix~\ref{app:memorisation-audit}.
\section{Test-time Ablations}\label{sec:ablations}
We study how sensitive the trained model is to changing the cycle pattern at inference time.
The question is prompted by the residual-stream dynamics in Appendix~\ref{app:resid}: a similar profile of large writes and rotations recurs with every module application, peaking near module boundaries. If each application performed an interchangeable refinement step, the number of applications could be changed without retraining. We test whether this holds for the $L$- and $H$-modules separately.
\paragraph{Setup.}
We evaluate Mimir v1 under varying recurrence patterns by changing the number of $H$ and $L$ cycles at inference time, keeping the weights fixed.
We vary $H \in \{1,2,3\}$ and $L \in \{1,\dots,5\}$, giving a recurrent depth $D = H(L{+}1)$ between $4$ and $12$, and evaluate on three single-token multiple-choice tasks (ARC-C, MMLU, HellaSwag) and two free-generation tasks (GSM8K, GEC-DaLA). Prompting and scoring follow the main evaluation setup (Section~\ref{sec:evals}), but all seven schedules are served by the same vLLM runtime with a different attention backend, so absolute values differ slightly from Tables~\ref{tab:english_summary}--\ref{tab:danish_summary}.
\paragraph{Results.}
Table~\ref{tab:testtime} shows that the trained schedule is best or near-best on every benchmark, so additional recurrent computation at inference does not improve accuracy. Varying the number of $L$-module applications degrades performance gracefully, whereas varying the number of $H$-module applications breaks the model, even at matched depth, where only the allocation of recurrent computation differs ($D{=}4$: $H{=}2$ vs.\ $H{=}1$; $D{=}12$: $H{=}2$ vs.\ $H{=}3$). A two-sided exact McNemar test on paired per-item GEC-DaLA outcomes ($N{=}1{,}024$) confirms both contrasts, with no discordant pairs in the failing direction ($672$ vs.\ $0$ and $816$ vs.\ $0$; $p < 10^{-200}$). Depth alone therefore does not predict whether the model works, and similar-looking residual dynamics do not imply interchangeable computation. Since we vary a single checkpoint, this shows that the model is specialised to its training-time $H$, not that $H{=}2$ is intrinsically required.
The two failing configurations fail in different ways: $H{=}1$ collapses on every task, falling below chance even on multiple choice. $H{=}3$ still emits well-formed option letters (at most $0.9\%$ invalid outputs) and loses only $7.1$ points on multiple choice on average, yet loses $85.7$ points on free generation, failing to produce a parseable answer on $99.3\%$ of GSM8K problems. Evaluating on multiple choice alone would thus understate the damage roughly twelvefold.
\begin{table}[t]
\centering
\small
\caption{Test-time recurrence sweep. The weights are fixed throughout; only
the inference schedule changes.
The shaded row is the schedule the model was trained with.
Best per column in \textbf{bold}.
$^\dagger$ Collapsed configuration: score below chance or near zero.}
\label{tab:testtime}
\begin{tabular}{rrrlccccc}
\toprule
$D$ & $H$ & $L$ & Schedule & ARC-C & MMLU & GSM8K & HellaSwag & GEC-DaLA \\
\midrule
4  & 1 & 3 & \texttt{LLLH}           & 11.09$^\dagger$ & 6.10$^\dagger$ & 0.00$^\dagger$ & 1.61$^\dagger$ & 0.00$^\dagger$ \\
4  & 2 & 1 & \texttt{LHLH}           & 37.29 & 38.58 & 46.63 & 33.30 & 65.63 \\
6  & 2 & 2 & \texttt{LLHLLH}         & 70.82 & 53.99 & 81.12 & 57.96 & 84.57 \\
\rowcolor{gray!25}
8  & 2 & 3 & \texttt{LLLHLLLH}       & 79.95 & 59.13 & \textbf{86.73} & \textbf{66.72} & \textbf{85.94} \\
10 & 2 & 4 & \texttt{LLLLHLLLLH}     & 79.69 & 58.74 & 84.76 & 60.94 & 85.16 \\
12 & 2 & 5 & \texttt{LLLLLHLLLLLH}   & \textbf{80.38} & \textbf{59.28} & 84.38 & 63.83 & 80.37 \\
12 & 3 & 3 & \texttt{LLLHLLLHLLLH}   & 74.83 & 52.35 & 0.53$^\dagger$ & 57.24 & 0.68$^\dagger$ \\
\bottomrule
\end{tabular}
\end{table}
\section{Related Work}
\paragraph{Recurrent Depth.}
\citet{dehghani2019universaltransformers} introduced the Universal Transformer, which reuses a shared block of weights across depth to decouple computation from parameter count.
\citet{yang2024looped} showed that looped transformers match standard transformers on a range of in-context learning problems while using less than 10\% of the parameters, and follow-up work has shown the approach's feasibility for language modeling~\citep{geiping2026scaling}. Recent looped language models include Ouro~\citep{zhu2025scaling}, trained on 7.7T tokens at 1.4B and 2.6B parameters, which we use as baselines.
\citet{ICLR2025_2676109d} found that performance depends on effective depth. \citet{kohli2026loop} show that recurrent models succeed at systematic compositional generalization.
Hierarchical reasoning models~\citep{wang2025hierarchicalreasoningmodel}, which introduced the L- and H-module split, were first proposed for algorithmic and puzzle reasoning.
HRM-Text~\citep{wang2026hrm} then showed that such models can be feasible as generalist language models, outperforming scale-matched looped transformers and standard transformers without weight sharing.
On these puzzle tasks, the Tiny Recursive Model~\citep{jolicoeur2025less} shows that the hierarchy is not needed: a single 7M-parameter, two-layer network that recursively refines its latent state outperforms HRM.
\paragraph{Synthetic data and transplant approaches.}
Synthetic datasets have emerged as a strong off-policy distillation strategy. Self-Instruct~\citep{wang2023self} introduced the core generate-and-filter paradigm by bootstrapping a model from its own generations. Alpaca-style pipelines further showed that strong instruction-following behavior can be induced from synthetic teacher outputs~\citep{taori2023alpaca}. Magpie~\citep{xu2025magpie} extended this direction with a pipeline focused on alignment-data synthesis, while also reporting strong instruction-tuning performance with substantially fewer examples than many larger alternatives. Our pipeline builds directly on this line of work: we use staged synthetic generation with quality gates before promoting samples into the training corpus.
A complementary approach grounds synthetic examples in given sources, such as existing datasets or raw text.
FineInstructions~\citep{patel2026fineinstructionsscalingsyntheticinstructions} is an example of this direction, scaling synthetic instruction pipelines to pre-training scale. We follow this approach in our transplant strategy, creating datasets with similar capabilities while respecting data rights and replacing faulty datasets when needed.
\paragraph{European efforts on low/medium-resource language modeling.}
A wide range of initiatives, especially in Europe, are currently building language models with a focus on their local languages, such as Apertus~\citep{apertus2025apertusdemocratizingopencompliant}, GPT-NL~\citep{van-oort-etal-2026-gpt}, NorwAI~\citep{gulla2026norwai} and AI Sweden~\citep{ekgren2024gpt}, as well as broader efforts like OpenEuroLLM~\citep{openeurollm2026scaling}. However, most of these efforts operate against the obstacle of collecting and training on massive pretraining corpora for LLM architectures.
European data-protection and copyright legislation raises both the cost and complexity of data acquisition and reuse. As a result, many projects train either relatively small models from scratch, limiting their capabilities, or adapt existing foundation models through continued pre- and post-training, which improves performance but does not deliver on full data sovereignty.
Compared to the conceptually most similar models, Mimir v1.5 attains higher domain averages than HRM-Text~1B, Ouro~1.4B, and Ouro~2.6B while training on roughly 10\% of Ouro's 7.7T tokens. One plausible contributor is our instruction-formatted training data, although we do not isolate this factor. Mimir~v1.5 thus shows that competitive sub-2B language models can be trained from scratch on permissible data only. Models, code, and all data except the two agreement-supplied sources will be made publicly available.
\section{Conclusion}
In this work, we presented Mimir v1 and v1.5, two sub-2B-parameter language models that leverage the HRM-Text architecture to provide highly competitive performance using only permissible data, excluding data containing personal information or copyright infringement and including data that is either openly licensed, made available by agreement, or allowed by the European Union's text and data mining exception for research institutions. We curated a diverse corpus, Mimir Training Corpus, consisting of 70.5B (v1) and more than 100B tokens per epoch (v1.5), including synthetic instruction datasets. We have shown that strong instruction-following and reasoning capabilities can be achieved without web-scale pre-training corpora or non-permissible data sources.
Mimir improves substantially over HRM-Text and Ouro, and is competitive with, or superior to, much larger models on several English and Danish benchmarks. It still lags behind Gemma 4 (5B, effective 2.3B) on the Math \& Code domains, making room for improvement in future iterations.
\paragraph{Limitations}
While Mimir demonstrated strong performance across all domains, punching above its weight class, several limitations apply.
First, its capabilities are shaped by the domains represented in the training mixture, which covers primarily English and Danish language tasks, mathematics, coding, and factual question answering. Because Mimir is not pretrained on broad web corpora, its performance may be limited on domains, styles, or factual areas that are weakly represented or absent from the mixture, more so than for conventionally pretrained models, which acquire broad world knowledge through web-scale next-token prediction.
Mimir is not multilingual, and while its capabilities may generalize to understand other languages, it is a bilingual model trained with focus on English and Danish. A further limitation concerns the amount and type of information that can be encoded in the relatively small model parameters. Even though Mimir exhibits good performance, factual knowledge is limited. Last, Mimir's inference time measurements scale, as expected based on effective depth at inference time, making it achieve throughput similar to a conventional non-looped LLM with an equivalent number of layers.
Future work will focus on investigating scaling behavior of HRM models like Mimir and reinforcement learning, which is yet unexplored for this architecture. Lastly, we will continue developing the dataset to test the limits of synthetic instruction data generation.
\subsection*{Ethics statement}
Training memory-efficient architectures contributes to the democratization of AI research, enabling practitioners to inference on consumer hardware.
In principle, open-source models come with risks of misuse. However, we believe that those risks are marginal given the small size of our model. To quantify copyright infringement risks, we ran two memorisation audits independently from each other, and found that only a negligible fraction of training data can be reproduced. Upon qualitative inspection, many of those were mathematical formulas and a popular children's song.
\subsection*{AI use statement}
In this work, we used generative AI tools to generate synthetic datasets, implement methods and experiments,  clean and reformat datasets, and provide feedback on research methodology or experiments (mock review). We have not used generative AI tools to interpret results. Development of theoretical models or conceptual frameworks, formulation of mathematical claims or critical ingredients for proving mathematical claims, proofs writing and hypotheses proposal or refinement are not applicable to this work. We have reviewed all AI-assisted work. LLM-generated evaluation code was verified and tested for correctness by 2 authors. We take responsibility for the final content of this work, including text, claims or artifacts produced with the aid of generative AI.
\subsection*{Reproducibility statement}
Both models, Mimir v1 and Mimir v1.5, the collection of synthetic training datasets, and the training and evaluation code, will be made publicly available. Evaluations were implemented and run in InspectAI~\citep{UK_AI_Security_Institute_Inspect_AI_Framework_2024} and are based on publicly available benchmark datasets. Details of the evaluation methodology are provided in Appendix~\ref{app:evals}. Architecture and training hyperparameters are listed in Appendix~\ref{app:hparams}.
\subsubsection*{Author Contributions}
We list contributions according to the Contributor Roles Taxonomy (CRediT)\footnote{\url{https://credit.niso.org}}:
\begin{description}
    \item[Peter Schneider-Kamp] \emph{Conceptualization}, \emph{Data curation}, \emph{Formal analysis}, \emph{Funding acquisition}, \emph{Investigation}, \emph{Methodology}, \emph{Project administration}, \emph{Resources}, \emph{Software},  \emph{Supervision}, \emph{Validation}, \emph{Visualization}, \emph{Writing -- original draft}, \emph{Writing -- review \& editing}
    \item[Jacob Nielsen] \emph{Formal analysis}, \emph{Investigation},
    \emph{Methodology},
    \emph{Software}, \emph{Validation}, \emph{Visualization}, \emph{Writing -- original draft}, \emph{Writing -- review \& editing}
    \item[Gianluca Barmina] \emph{Data curation}, \emph{Investigation},  \& \emph{Writing -- review \& editing}
    \item[Danial Namazifard] \emph{Investigation}, \emph{Software} \emph{Writing -- review \& editing}
    \item[Annemette Brok Pirchert] \emph{Formal analysis}, \emph{Investigation}, \emph{Software}, \emph{Writing -- review \& editing}, \emph{Visualization}
    \item[Kenneth Enevoldsen] \emph{Data curation}, \emph{Resources}, \emph{Writing -- review \& editing}
    \item[Lukas Galke Poech]  \emph{Formal analysis}, \emph{Investigation},
    \emph{Methodology},
    \emph{Software}, \emph{Supervision},
    \emph{Validation}, \emph{Visualization},
     \emph{Writing -- original draft}, \emph{Writing -- review \& editing}
\end{description}
\paragraph{Additional Contributors}
\begin{description}
    \item[Mogens Henrik From]  \emph{Investigation}, \emph{Project administration}, \emph{Software}
    \item[Andrea Blasi Núñez]  \emph{Formal analysis}, \emph{Software}, \emph{Visualization}
    \item[Stine Lyngsø Beltoft] \emph{Data curation}, \emph{Resources}
    \item[Torben Blach] \emph{Project administration}, \emph{Funding acquisition}
    \item[Sofie Helene Bruun] \emph{Data curation}, \emph{Resources}
    \item[Oliver Kinch] \emph{Data curation}, \emph{Resources}
    \item[Rasmus Larsen] \emph{Data curation}, \emph{Resources}
    \item[Dan Saattrup Smart] \emph{Data curation}, \emph{Resources}
    \item[Kristoffer Laigaard Nielbo] \emph{Funding acquisition}
\end{description}
\subsubsection*{Acknowledgments}
The research was supported by the Danish Foundation Models project, funded by the Ministry of Science, Higher
Education and Digital Affairs.
D.N., A.B.P., and L.G.P.\ further acknowledge support from the MIST project, funded by the Novo Nordisk Foundation under grant reference number NNF25OC0103204.
The computation for this project was performed on the UCloud interactive HPC system managed by the eScience Center at the University of Southern Denmark.
\bibliography{iclr2027_conference}

@article{jolicoeur2025less,
  title={Less is more: Recursive reasoning with tiny networks},
  author={Jolicoeur-Martineau, Alexia},
  journal={arXiv preprint arXiv:2510.04871},
  year={2025}
}

@inproceedings{ICLR2025_2676109d,
 author = {Saunshi, Nikunj and Dikkala, Nishanth and Li, Zhiyuan and Kumar, Sanjiv and J. Reddi, Sashank},
 booktitle = {International Conference on Learning Representations},
 editor = {Y. Yue and A. Garg and N. Peng and F. Sha and R. Yu},
 pages = {14855--14881},
 title = {Reasoning with Latent Thoughts: On the Power of Looped Transformers},
 url = {https://proceedings.iclr.cc/paper_files/paper/2025/file/2676109d49d1eb26d6bc584a8f556305-Paper-Conference.pdf},
 volume = {2025},
 year = {2025}
}

@misc{openeurollm2026scaling,
      title={Deriving Scaling Laws for OpenEuroLLM Models: Learning Rate, Batch Size and Loss}, 
      author={{OpenEuroLLM Consortium}},
      year={2026},
      eprint={2608.28308},
      archivePrefix={arXiv},
      primaryClass={cs.CL}
}

@inproceedings{ekgren2024gpt,
  title={GPT-SW3: An autoregressive language model for the Scandinavian languages},
  author={Ekgren, Ariel and Gyllensten, Amaru Cuba and Stollenwerk, Felix and {\"O}hman, Joey and Isbister, Tim and Gogoulou, Evangelia and Carlsson, Fredrik and Casademont, Judit and Sahlgren, Magnus},
  booktitle={Proceedings of the 2024 Joint International Conference on Computational Linguistics, Language Resources and Evaluation (LREC-COLING 2024)},
  pages={7886--7900},
  year={2024}
}

@article{gulla2026norwai,
  title={{NorwAI}'s Large Language Models: Technical Report},
  author={Gulla, Jon Atle and Liu, Peng and Zhang, Lemei},
  journal={arXiv preprint arXiv:2601.03034},
  year={2026}
}

@inproceedings{van-oort-etal-2026-gpt,
    title = "{GPT}-{NL} Public Corpus: A Permissively Licensed, {D}utch-First Dataset for {LLM} Pre-training",
    author = "Van Oort, Jesse J.  and
      Brinkkemper, Frank  and
      de Graaf, Erik  and
      Vanroy, Bram  and
      Lensink, Saskia",
    editor = "Piperidis, Stelios  and
      Bel, N{\'u}ria  and
      van den Heuvel, Henk  and
      Ide, Nancy  and
      Krek, Simon  and
      Toral, Antonio",
    booktitle = "Proceedings of the Fifteenth Language Resources and Evaluation Conference",
    month = may,
    year = "2026",
    address = "Palma de Mallorca, Spain",
    publisher = "ELRA Language Resource Association",
    url = "https://aclanthology.org/2026.lrec-1.548/",
    doi = "10.63317/5fbtc336wwx2",
    pages = "6893--6903"
}

@misc{apertus2025apertusdemocratizingopencompliant,
      title={Apertus: Democratizing Open and Compliant {LLMs} for Global Language Environments}, 
      author={Project Apertus and Alejandro Hernández-Cano and Alexander Hägele and Allen Hao Huang and Angelika Romanou and Antoni-Joan Solergibert and Barna Pasztor and Bettina Messmer and Dhia Garbaya and Eduard Frank Ďurech and Ido Hakimi and Juan García Giraldo and Mete Ismayilzada and Negar Foroutan and Skander Moalla and Tiancheng Chen and Vinko Sabolčec and Yixuan Xu and Michael Aerni and Badr AlKhamissi and Inés Altemir Mariñas and Mohammad Hossein Amani and Matin Ansaripour and Ilia Badanin and Harold Benoit and Emanuela Boros and Nicholas Browning and Fabian Bösch and Maximilian Böther and Niklas Canova and Camille Challier and Clement Charmillot and Jonathan Coles and Jan Deriu and Arnout Devos and Lukas Drescher and Daniil Dzenhaliou and Maud Ehrmann and Dongyang Fan and Simin Fan and Silin Gao and Miguel Gila and María Grandury and Diba Hashemi and Alexander Hoyle and Jiaming Jiang and Mark Klein and Andrei Kucharavy and Anastasiia Kucherenko and Frederike Lübeck and Roman Machacek and Theofilos Manitaras and Andreas Marfurt and Kyle Matoba and Simon Matrenok and Henrique Mendonça and Fawzi Roberto Mohamed and Syrielle Montariol and Luca Mouchel and Sven Najem-Meyer and Jingwei Ni and Gennaro Oliva and Matteo Pagliardini and Elia Palme and Andrei Panferov and Léo Paoletti and Marco Passerini and Ivan Pavlov and Auguste Poiroux and Kaustubh Ponkshe and Nathan Ranchin and Javi Rando and Mathieu Sauser and Jakhongir Saydaliev and Muhammad Ali Sayfiddinov and Marian Schneider and Stefano Schuppli and Marco Scialanga and Andrei Semenov and Kumar Shridhar and Raghav Singhal and Anna Sotnikova and Alexander Sternfeld and Ayush Kumar Tarun and Paul Teiletche and Jannis Vamvas and Xiaozhe Yao and Hao Zhao and Alexander Ilic and Ana Klimovic and Andreas Krause and Caglar Gulcehre and David Rosenthal and Elliott Ash and Florian Tramèr and Joost VandeVondele and Livio Veraldi and Martin Rajman and Thomas Schulthess and Torsten Hoefler and Antoine Bosselut and Martin Jaggi and Imanol Schlag},
      year={2025},
      eprint={2509.14233},
      archivePrefix={arXiv},
      primaryClass={cs.CL},
      url={https://arxiv.org/abs/2509.14233}, 
}

@misc{patel2026fineinstructionsscalingsyntheticinstructions,
      title={FineInstructions: Scaling Synthetic Instructions to Pre-Training Scale}, 
      author={Ajay Patel and Colin Raffel and Chris Callison-Burch},
      year={2026},
      eprint={2601.22146},
      archivePrefix={arXiv},
      primaryClass={cs.CL},
      url={https://arxiv.org/abs/2601.22146}, 
}

@inproceedings{xu2025magpie,
  title={Magpie: Alignment data synthesis from scratch by prompting aligned {LLMs} with nothing},
  author={Xu, Zhangchen and Jiang, Fengqing and Niu, Luyao and Deng, Yuntian and Poovendran, Radha and Choi, Yejin and Lin, Bill Yuchen},
  booktitle={International Conference on Learning Representations},
  volume={2025},
  pages={76346--76382},
  year={2025}
}

@article{taori2023alpaca,
  title={Alpaca: A strong, replicable instruction-following model},
  author={Taori, Rohan and Gulrajani, Ishaan and Zhang, Tianyi and Dubois, Yann and Li, Xuechen and Guestrin, Carlos and Liang, Percy and Hashimoto, Tatsunori B},
  journal={Stanford Center for Research on Foundation Models. https://crfm. stanford. edu/2023/03/13/alpaca. html},
  volume={3},
  number={6},
  pages={7},
  year={2023}
}

@inproceedings{wang2023self,
  title={Self-instruct: Aligning language models with self-generated instructions},
  author={Wang, Yizhong and Kordi, Yeganeh and Mishra, Swaroop and Liu, Alisa and Smith, Noah A and Khashabi, Daniel and Hajishirzi, Hannaneh},
  booktitle={Proceedings of the 61st annual meeting of the association for computational linguistics (volume 1: long papers)},
  pages={13484--13508},
  year={2023}
}

@inproceedings{yang2024looped,
  title={Looped transformers are better at learning learning algorithms},
  author={Yang, Liu and Lee, Kangwook and Nowak, Robert and Papailiopoulos, Dimitris},
  booktitle={International conference on learning representations},
  volume={2024},
  pages={42195--42214},
  year={2024}
}

@misc{dehghani2019universaltransformers,
      title={Universal Transformers}, 
      author={Mostafa Dehghani and Stephan Gouws and Oriol Vinyals and Jakob Uszkoreit and Łukasz Kaiser},
      year={2019},
      eprint={1807.03819},
      archivePrefix={arXiv},
      primaryClass={cs.CL},
      url={https://arxiv.org/abs/1807.03819}, 
}

@article{su2024roformer,
  title={Roformer: Enhanced transformer with rotary position embedding},
  author={Su, Jianlin and Ahmed, Murtadha and Lu, Yu and Pan, Shengfeng and Bo, Wen and Liu, Yunfeng},
  journal={Neurocomputing},
  volume={568},
  pages={127063},
  year={2024},
  publisher={Elsevier}
}

@article{wolf2019huggingface,
  title={Huggingface's transformers: State-of-the-art natural language processing},
  author={Wolf, Thomas and Debut, Lysandre and Sanh, Victor and Chaumond, Julien and Delangue, Clement and Moi, Anthony and Cistac, Pierric and Rault, Tim and Louf, R{\'e}mi and Funtowicz, Morgan and others},
  journal={arXiv preprint arXiv:1910.03771},
  year={2019}
}

@article{zadouri2026flashattention,
  title={Flashattention-4: Algorithm and kernel pipelining co-design for asymmetric hardware scaling},
  author={Zadouri, Ted and Hoehnerbach, Markus and Shah, Jay and Thakkar, Vijay and Dao, Tri},
  journal={Proceedings of Machine Learning and Systems},
  volume={8},
  pages={912--926},
  year={2026}
}

@article{liu2025acereason,
  title={AceReason-Nemotron 1.1: Advancing Math and Code Reasoning through SFT and RL Synergy},
  author={Liu, Zihan and Yang, Zhuolin and Chen, Yang and Lee, Chankyu and Shoeybi, Mohammad and Catanzaro, Bryan and Ping, Wei},
  journal={arXiv preprint arXiv:2506.13284},
  year={2025}
}

@article{toshniwal2024openmath2,
  title   = {OpenMathInstruct-2: Accelerating AI for Math with Massive Open-Source Instruction Data},
  author  = {Shubham Toshniwal and Wei Du and Ivan Moshkov and  Branislav Kisacanin and Alexan Ayrapetyan and Igor Gitman},
  year    = {2024},
  journal = {arXiv preprint arXiv:2410.01560}
}

@inproceedings{kwon2023efficient,
  title={Efficient Memory Management for Large Language Model Serving with PagedAttention},
  author={Woosuk Kwon and Zhuohan Li and Siyuan Zhuang and Ying Sheng and Lianmin Zheng and Cody Hao Yu and Joseph E. Gonzalez and Hao Zhang and Ion Stoica},
  booktitle={Proceedings of the ACM SIGOPS 29th Symposium on Operating Systems Principles},
  year={2023}
}

@misc{gemmateam2026gemma4technicalreport,
      title={Gemma 4 Technical Report}, 
      author={{Gemma Team}},
      year={2026},
      eprint={2607.02770},
      archivePrefix={arXiv},
      primaryClass={cs.CL},
      url={https://arxiv.org/abs/2607.02770}, 
}

@inproceedings{loshchilov2017decoupled,
  title={Decoupled weight decay regularization},
  author={Loshchilov, Ilya and Hutter, Frank},
  booktitle={ICLR},
  year={2019}
}

@article{clark2018think,
  title={Think you have solved question answering? {T}ry {ARC}, the {Ai}2 reasoning challenge},
  author={Clark, Peter and Cowhey, Isaac and Etzioni, Oren and Khot, Tushar and Sabharwal, Ashish and Schoenick, Carissa and Tafjord, Oyvind},
  journal={arXiv preprint arXiv:1803.05457},
  year={2018}
}

@misc{UK_AI_Security_Institute_Inspect_AI_Framework_2024,
  author = {AI Security Institute, UK},
  title = {Inspect {AI:} {Framework} for {Large} {Language} {Model}
    {Evaluations}},
  date = {2024-05},
  year= {2024},
  url = {https://github.com/UKGovernmentBEIS/inspect_ai},
  langid = {en}
}

@misc{6c685239843b47bfb7f47a78aae67929,
title = "Dynaword: From One-shot to Continuously Developed Datasets",
author = "Kenneth Enevoldsen and Jensen, Kristian N{\o}rgaard and Jan Kostkan and Bal{\'a}zs Szab{\'o} and M{\'a}rton Kardos and Kirsten Vad and Johan Heinsen and N{\'u}{\~n}ez, Andrea Blasi and Gianluca Barmina and Jacob Nielsen and Rasmus Larsen and Peter Vahlstrup and Dalum, Per M{\o}ldrup and Desmond Elliott and Lukas Galke and Peter Schneider-Kamp and Kristoffer Nielbo",
year = "2025",
doi = "10.48550/arXiv.2508.02271",
language = "English",
publisher = "arXiv",
type = "WorkingPaper",
institution = "arXiv",
}

@misc{kandpal2025commonpilev018tb,
      title={The Common Pile v0.1: An 8TB Dataset of Public Domain and Openly Licensed Text}, 
      author={Nikhil Kandpal and Brian Lester and Colin Raffel and Sebastian Majstorovic and Stella Biderman and Baber Abbasi and Luca Soldaini and Enrico Shippole and A. Feder Cooper and Aviya Skowron and John Kirchenbauer and Shayne Longpre and Lintang Sutawika and Alon Albalak and Zhenlin Xu and Guilherme Penedo and Loubna Ben Allal and Elie Bakouch and John David Pressman and Honglu Fan and Dashiell Stander and Guangyu Song and Aaron Gokaslan and Tom Goldstein and Brian R. Bartoldson and Bhavya Kailkhura and Tyler Murray},
      year={2025},
      eprint={2506.05209},
      archivePrefix={arXiv},
      primaryClass={cs.CL},
      url={https://arxiv.org/abs/2506.05209}, 
}

@article{wang2026hrm,
  title={HRM-Text: Efficient Pretraining Beyond Scaling},
  author={Wang, Guan and Liu, Changling and Wang, Chenyu and Zhou, Cai and Sun, Yuhao and Wu, Yifei and Zhen, Shuai and Scimeca, Luca and Yadkori, Yasin Abbasi},
  journal={arXiv preprint arXiv:2605.20613},
  year={2026}
}

@misc{olmo2025olmo3,
title={Olmo 3},
author={{Team Olmo}},
year={2025},
eprint={2512.13961},
archivePrefix={arXiv},
primaryClass={cs.CL},
url={https://arxiv.org/abs/2512.13961},
}

@article{lambert2024tulu,
  title={Tulu 3: Pushing frontiers in open language model post-training},
  author={Lambert, Nathan and Morrison, Jacob and Pyatkin, Valentina and Huang, Shengyi and Ivison, Hamish and Brahman, Faeze and Miranda, Lester James V and Liu, Alisa and Dziri, Nouha and Lyu, Shane and others},
  journal={arXiv preprint arXiv:2411.15124},
  year={2024}
}

@misc{propme,
      title={{LLMs} Can Leak Training Data But Do They Want To? A Propensity-Aware Evaluation of Memorization in {LLMs}}, 
      author={Gianluca Barmina and Peter Schneider-Kamp and Lukas Galke Poech},
      year={2026},
      eprint={2606.06286},
      archivePrefix={arXiv},
      primaryClass={cs.CL},
      url={https://arxiv.org/abs/2606.06286}, 
}

@inproceedings{giannou2023looped,
  title={Looped transformers as programmable computers},
  author={Giannou, Angeliki and Rajput, Shashank and Sohn, Jy-yong and Lee, Kangwook and Lee, Jason D and Papailiopoulos, Dimitris},
  booktitle={International Conference on Machine Learning},
  pages={11398--11442},
  year={2023},
  organization={PMLR}
}

@book{kahneman2011thinking,
  title={Thinking, fast and slow},
  author={Kahneman, Daniel},
  year={2011},
  publisher={Macmillan}
}

@misc{wang2025hierarchicalreasoningmodel,
      title={Hierarchical Reasoning Model}, 
      author={Guan Wang and Jin Li and Yuhao Sun and Xing Chen and Changling Liu and Yue Wu and Meng Lu and Sen Song and Yasin Abbasi Yadkori},
      year={2025},
      eprint={2506.21734},
      archivePrefix={arXiv},
      primaryClass={cs.AI},
      url={https://arxiv.org/abs/2506.21734}, 
}

@article{slack-al-moubayed-2025-early,
    title = "Early Detection and Reduction of Memorization for Domain Adaptation and Instruction Tuning",
    author = "Slack, Dean L.  and
      Al Moubayed, Noura",
    journal = "Transactions of the Association for Computational Linguistics",
    volume = "13",
    year = "2025",
    address = "Cambridge, MA",
    publisher = "MIT Press",
    url = "https://aclanthology.org/2025.tacl-1.66/",
    doi = "10.1162/tacl.a.49",
    pages = "1459--1473"
}

@misc{OpenHermes2.5,
  title = {OpenHermes 2.5: An Open Dataset of Synthetic Data for Generalist LLM Assistants},
  author = {Teknium},
  year = {2023},
  publisher = {HuggingFace},
  url = {https://huggingface.co/datasets/teknium/OpenHermes-2.5}
}

@article{wang2025slmsurvey,
author = {Wang, Fali and Zhang, Zhiwei and Zhang, Xianren and Wu, Zongyu and Mo, TzuHao and Lu, Qiuhao and Wang, Wanjing and Li, Rui and Xu, Junjie and Tang, Xianfeng and He, Qi and Ma, Yao and Huang, Ming and Wang, Suhang},
title = {A Comprehensive Survey of Small Language Models in the Era of Large Language Models: Techniques, Enhancements, Applications, Collaboration with {LLMs}, and Trustworthiness},
year = {2025},
issue_date = {December 2025},
publisher = {Association for Computing Machinery},
address = {New York, NY, USA},
volume = {16},
number = {6},
issn = {2157-6904},
url = {https://doi.org/10.1145/3768165},
doi = {10.1145/3768165},
journal = {ACM Trans. Intell. Syst. Technol.},
month = nov,
articleno = {145},
numpages = {87}
}

@inproceedings{kandpal2022deduplicating,
  title={Deduplicating training data mitigates privacy risks in language models},
  author={Kandpal, Nikhil and Wallace, Eric and Raffel, Colin},
  booktitle={International Conference on Machine Learning},
  pages={10697--10707},
  year={2022},
  organization={PMLR}
}

@inproceedings{
carlini2023quantifying,
title={Quantifying Memorization Across Neural Language Models},
author={Nicholas Carlini and Daphne Ippolito and Matthew Jagielski and Katherine Lee and Florian Tramer and Chiyuan Zhang},
booktitle={The Eleventh International Conference on Learning Representations },
year={2023},
url={https://openreview.net/forum?id=TatRHT_1cK}
}

@article{zhu2025scaling,
  title={Scaling latent reasoning via looped language models},
  author={Zhu, Rui-Jie and Wang, Zixuan and Hua, Kai and Zhang, Tianyu and Li, Ziniu and Que, Haoran and Wei, Boyi and Wen, Zixin and Yin, Fan and Xing, He and others},
  journal={arXiv preprint arXiv:2510.25741},
  year={2025}
}

@inproceedings{lee-etal-2022-deduplicating,
    title = "Deduplicating Training Data Makes Language Models Better",
    author = "Lee, Katherine  and
      Ippolito, Daphne  and
      Nystrom, Andrew  and
      Zhang, Chiyuan  and
      Eck, Douglas  and
      Callison-Burch, Chris  and
      Carlini, Nicholas",
    no_editor = "Muresan, Smaranda  and
      Nakov, Preslav  and
      Villavicencio, Aline",
    booktitle = "Proceedings of the 60th Annual Meeting of the Association for Computational Linguistics (Volume 1: Long Papers)",
    month = may,
    year = "2022",
    address = "Dublin, Ireland",
    publisher = "Association for Computational Linguistics",
    url = "https://aclanthology.org/2022.acl-long.577/",
    doi = "10.18653/v1/2022.acl-long.577",
    pages = "8424--8445",
}

@inproceedings{clark2019boolq,
  title     = {{BoolQ}: Exploring the Surprising Difficulty of Natural Yes/No Questions},
  author    = {Clark, Christopher and Lee, Kenton and Chang, Ming-Wei and Kwiatkowski, Tom and Collins, Michael and Toutanova, Kristina},
  booktitle = {Proceedings of the 2019 Conference of the North {A}merican Chapter of the Association for Computational Linguistics: Human Language Technologies, Volume 1 (Long and Short Papers)},
  month     = {jun},
  year      = {2019},
  address   = {Minneapolis, Minnesota},
  publisher = {Association for Computational Linguistics},
  pages     = {2924--2936},
  url       = {https://aclanthology.org/N19-1300/}
}

@article{winogrande,
author = {Sakaguchi, Keisuke and Bras, Ronan Le and Bhagavatula, Chandra and Choi, Yejin},
title = {WinoGrande: an adversarial winograd schema challenge at scale},
year = {2021},
issue_date = {September 2021},
publisher = {Association for Computing Machinery},
address = {New York, NY, USA},
volume = {64},
number = {9},
issn = {0001-0782},
url = {https://doi.org/10.1145/3474381},
doi = {10.1145/3474381},
journal = {Commun. ACM},
month = aug,
pages = {99–106},
numpages = {8}
}

@inproceedings{zellers-etal-2019-hellaswag,
    title = "{H}ella{S}wag: Can a Machine Really Finish Your Sentence?",
    author = "Zellers, Rowan  and
      Holtzman, Ari  and
      Bisk, Yonatan  and
      Farhadi, Ali  and
      Choi, Yejin",
    editor = "Korhonen, Anna  and
      Traum, David  and
      M{\`a}rquez, Llu{\'i}s",
    booktitle = "Proceedings of the 57th Annual Meeting of the Association for Computational Linguistics",
    month = jul,
    year = "2019",
    address = "Florence, Italy",
    publisher = "Association for Computational Linguistics",
    url = "https://aclanthology.org/P19-1472/",
    doi = "10.18653/v1/P19-1472",
    pages = "4791--4800"
}

@misc{hendrycks2021mmlu,
      title={Measuring Massive Multitask Language Understanding}, 
      author={Dan Hendrycks and Collin Burns and Steven Basart and Andy Zou and Mantas Mazeika and Dawn Song and Jacob Steinhardt},
      year={2021},
      eprint={2009.03300},
      archivePrefix={arXiv},
      primaryClass={cs.CY},
      url={https://arxiv.org/abs/2009.03300}, 
}

@misc{dua2019dropreadingcomprehensionbenchmark,
      title={DROP: A Reading Comprehension Benchmark Requiring Discrete Reasoning Over Paragraphs}, 
      author={Dheeru Dua and Yizhong Wang and Pradeep Dasigi and Gabriel Stanovsky and Sameer Singh and Matt Gardner},
      year={2019},
      eprint={1903.00161},
      archivePrefix={arXiv},
      primaryClass={cs.CL},
      url={https://arxiv.org/abs/1903.00161}, 
}

@misc{huang2021govreport,
      title={Efficient Attentions for Long Document Summarization}, 
      author={Luyang Huang and Shuyang Cao and Nikolaus Parulian and Heng Ji and Lu Wang},
      year={2021},
      eprint={2104.02112},
      archivePrefix={arXiv},
      primaryClass={cs.CL}
    }

@misc{cobbe2021gsm8k,
      title={Training Verifiers to Solve Math Word Problems}, 
      author={Karl Cobbe and Vineet Kosaraju and Mohammad Bavarian and Mark Chen and Heewoo Jun and Lukasz Kaiser and Matthias Plappert and Jerry Tworek and Jacob Hilton and Reiichiro Nakano and Christopher Hesse and John Schulman},
      year={2021},
      eprint={2110.14168},
      archivePrefix={arXiv},
      primaryClass={cs.LG},
      url={https://arxiv.org/abs/2110.14168}, 
}

@article{hendrycks2021math,
  title={Measuring Mathematical Problem Solving With the MATH Dataset},
  author={Dan Hendrycks and Collin Burns and Saurav Kadavath and Akul Arora and Steven Basart and Eric Tang and Dawn Song and Jacob Steinhardt},
  journal={NeurIPS},
  year={2021}
}

@article{chen2021humaneval,
  title={Evaluating Large Language Models Trained on Code},
  author={Mark Chen and Jerry Tworek and Heewoo Jun and Qiming Yuan and Henrique Ponde de Oliveira Pinto and Jared Kaplan and Harri Edwards and Yuri Burda and Nicholas Joseph and Greg Brockman and Alex Ray and Raul Puri and Gretchen Krueger and Michael Petrov and Heidy Khlaaf and Girish Sastry and Pamela Mishkin and Brooke Chan and Scott Gray and Nick Ryder and Mikhail Pavlov and Alethea Power and Lukasz Kaiser and Mohammad Bavarian and Clemens Winter and Philippe Tillet and Felipe Petroski Such and Dave Cummings and Matthias Plappert and Fotios Chantzis and Elizabeth Barnes and Ariel Herbert-Voss and William Hebgen Guss and Alex Nichol and Alex Paino and Nikolas Tezak and Jie Tang and Igor Babuschkin and Suchir Balaji and Shantanu Jain and William Saunders and Christopher Hesse and Andrew N. Carr and Jan Leike and Josh Achiam and Vedant Misra and Evan Morikawa and Alec Radford and Matthew Knight and Miles Brundage and Mira Murati and Katie Mayer and Peter Welinder and Bob McGrew and Dario Amodei and Sam McCandlish and Ilya Sutskever and Wojciech Zaremba},
  year={2021},
  journal={arXiv preprint 2107.03374},
}

@inproceedings{pauli2021danlp_angrytweets,
  title={DaNLP: An open-source toolkit for Danish Natural Language Processing},
  author={Pauli, Amalie Brogaard and Barrett, Maria and Lacroix, Oph{\'e}lie and Hvingelby, Rasmus},
  booktitle={Proceedings of the 23rd Nordic Conference on Computational Linguistics (NoDaLiDa)},
  pages={460--466},
  year={2021}
}

@inproceedings{barmina2026dala,
  title={{DaLA}: Danish linguistic acceptability evaluation guided by real world errors},
  author={Barmina, Gianluca and Norman, Nathalie Carmen Hau and Schneider-Kamp, Peter and Poech, Lukas Galke},
  booktitle={Proceedings of the Fifteenth Language Resources and Evaluation Conference (LREC 2026)},
  no_volume={11},
  no_number={16},
  pages={4312--4326},
  year={2026}
}

@misc{chang2026globalpiqaevaluatingcommonsense,
      title={Global PIQA: Evaluating Commonsense Reasoning Across 100+ Languages and Cultures}, 
      author={Tyler A. Chang and Catherine Arnett and Abdelrahman Sadallah and Abdelrahman Eldesokey and Abeer Kashar and Abolade Daud and Abosede Grace Olanihun and Adamu Labaran Mohammed and Adeyemi Praise and Adhikarimayum Meerajita Sharma and Aditi Gupta and Adril Putra Merin and Adwoa Bremang and Afitab Iyigun and Afonso Simplício and Ahmed Essouaied and Aicha Chorana and Akhil Eppa and Akintunde Oladipo and Akriti Kuri and Akshay Ramesh and Aleksei Dorkin and Alfred Malengo Kondoro and Alham Fikri Aji and Ali Eren Çetintaş and Allan Hanbury and Alou Dembele and Alp Niksarli and Álvaro Arroyo and Amin Bajand and Amol Khanna and Ana Chkhaidze and Ana Carolina Condez and Anamaria-Roberta Hartl and Andiswa Mkhonto and Andrew Hoblitzell and Andrew Tran and Angelos Poulis and Anirban Majumder and Anjali Chaudhary and Anna Vacalopoulou and Annette Kuuipolani Kanahele Wong and Annika Simonsen and Anton Kovalev and Anupam Nayak and Ashvanth S and Ayodeji Lana and Ayu Purwarianti and Bashar Alhafni and Benedict Busole and Bernard Ghanem and Bharti Nathani and Biljana Stojanovska Đurić and Blessing Ogundipe and Bolaotan Agbonile and Bragi Bergsson and Bruce Torres Fischer and Burak Tutar and Burcu Çınar and Cade Kane and Can Udomcharoenchaikit and Chadi Helwe and Chaithra Reddy Nerella and Chen Cecilia Liu and Chiamaka Nwokolo and Christopher Homan and Clément Sampebgo and Cristina España-Bonet and Cynthia Amol and Daeyoep Lee and Dan Saattrup Smart and Dana Arad and Daniil Dzenhaliou and Dasol Choi and David Liu and David Semedo and David Anugraha and Deborah Popoola and Deividas Mataciunas and Delphine Nyaboke and Dennis Owusu and Dhyuthy Krishna Kumar and Diogo Tavares and Diogo Glória-Silva and Divyanshu Goyal and DongGeon Lee and E. Kelly Buchanan and Ebele Nwamaka Anajemba and Egonu Ngozi Grace and Elena Mickel and Elias Herranen and Eliza Acharya and Eman Nisar and Emile Anand and Emmanuel Habumuremyi and Emuobonuvie Maria Ajiboye and Eryawan Presma Yulianrifat and Esther Adenuga and Ewa Rudnicka and Faith Itiola and Faran Taimoor Butt and Fareeha Fayyaz Sheikh and Fathima Thekkekara and Fatima Haouari and Faustin Nsengiyumva and Fenal Ashokbhai Ilasariya and Filbert Aurelian Tjiaranata and Firas Laakom and Francesca Grasso and Francesco Periti and Francesco Orabona and Gbenga Kayode Solomon and Genta Indra Winata and Gia Nghia Ngo and Gloria Udhedhe-oze and Gonçalo Vinagre and Gopi Naga Sai Ram Challagolla and Gorka Urbizu-Garmendia and Gouthami Vadithya and Guijin Son and Gulnaz Abdykadyrova and Gyan Swaroop Mohapatra and Hafeez Ullah and Hafsteinn Einarsson and Hai Hu and Hamidreza Saffari and Hamza Zaidi and Haopeng Zhang and Harethah Abu Shairah and Harry Vuong and Hele-Andra Kuulmets and Hitesh Laxmichand Patel and Houda Bouamor and Hwanjo Yu and Iben Nyholm Debess and İbrahim Ethem Deveci and Ikhlasul Akmal Hanif and Ikhyun Cho and Inês Vieira and Inês Calvo and Isaac Manzi and Ismael Illa Salifou and Ismail Daud and Ismail Yusuf and Itay Itzhak and Ivan Zhelyazkov and Ivan Belashkin and Ivan Spada and Jacob Brinton and Jafar Isbarov and Jaka Čibej and Jan Kocoń and Jan Cuhel and Jauza Krito and Jebish Purbey and Jennifer Za and Jennifer Mickel and Jenny Kunz and Jessica Ratovondranto and Jeyarajalingam Varsha and Jihae Jeong and Jimena Tena Dávalos and Jinu Lee and João Magalhães and John Seon Keun Yi and Jongin Kim and Joseph Chataignon and Joseph Marvin Imperial and Jubeerathan Thevakumar and Judith Land and Julia Alekseenko and Junchen Jiang and Jungwhan Kim and Kairit Sirts and Kamesh R and Kamesh V and Kanda Tshinu and Kätriin Kukk and Kaustubh Ponkshe and Kavsar Huseynova and Ke He and Kenneth Enevoldsen and Kent Joshua Alvarez and Kerem Zaman and Khalil Mrini and Kian Kyars and Komal Gour and Krishnakumar Lainitha and Krister Kruusmaa and Kunal Mukherjee and Kusum Chouhan and Laura Castro and Laura M. Porrino-Moscoso and Lenny Sivi Za Nzambi and Leshem Choshen and Levent Sencan and Lilja Øvrelid and Lisa Alazraki and Loretta Oma Jones and Lovina Ehimen-Ugbede and Luheerathan Thevakumar and Luxshan Thavarasa and Mahnoor Malik and Mamadou K. Keita and Mansi Jangid and Marco De Santis and Marcos Garcia and Marek Šuppa and Mariam D'Ciofalo and Marii Ojastu and Marium Attaullah and Maryam Sikander and Mausami Narayan and Maximos Skandalis and Mehak Mehak and Mehmet İlteriş Bozkurt and Melaku Bayu and Menan Velayuthan and Mhasilenuo Vizo and Michael Leventhal and Michał Marcińczuk and Mina Almasi and Mirna Potočnjak and Mithil Bangera and Mohammadamin Shafiei and Mohiba Ansari and Mridul Sharma and Mrityunjaya Indoria and Mughees Ur Rehman and Muhammad Ravi Shulthan Habibi and Murat Kolić and Murat Barkın Kınay and Nada Galant and Naina Singh Rathore and Naphat Permpredanun and Narada Maugin and Nathalie Norman and Nicholas Kluge Corrêa and Nikola Ljubešić and Nirmal Thomas and Nisansa de Silva and Nisheeth Joshi and Nitish Ponkshe and Nizar Habash and Nneoma Udeze and Noel Thomas and Noémi Ligeti-Nagy and Nouhoum Coulibaly and Odunayo Ogundepo and Odunayo Kareemat Buliaminu and Oghojafor Godswill Fejiro and Okechukwu God'spraise and Olanrewaju Samuel and Olaoye Deborah Oluwaseun and Olasoji Akindejoye and Olga Snissarenko and Onyinye Anulika Chiemezie and Orkun Kınay and Osman Tursun and Oyelade Oluwafemi Joshua and Oyesanmi Fiyinfoluwa and Pablo Rodríguez and Pablo Gamallo and Palak Arora and Pedro Valente and Peter Rupnik and Philip Oghenesuowho Ekiugbo and Prakhar Agarwal and Pramit Sahoo and Prokopis Prokopidis and Pua Niau-Puhipau and Quadri Yahya and Rachele Mignone and Raghav Singhal and Rahul Raja and Ram Mohan Rao Kadiyala and Raphael Merx and Rasmus Larsen and Ratnavel Rajalakshmi and Rishav Ghosh and Romina Oji and Ron Kekeha Solis and Rui Guerra and Rushikesh Zawar and Sa'ad Nasir Bashir and Saeed Alzaabi and Sahil Sandeep and Sai Pavan Batchu and Sai Sandeep Kantareddy and Saleha Muzammil and Salsabila Zahirah Pranida and Sam Buchanan and Samuel Rutunda and Sander Land and Sarah Sulollari and Sardar Ali and Saroj Sapkota and Sarveswaran Kengatharaiyer and Saulius Tautvaisas and Sayambhu Sen and Sayantani Banerjee and Sebastien Diarra and Segun Afolayan and Senthilnathan M and Sewoong Lee and Shaan Shah and Shankar Venkitachalam and Sharifa Djurabaeva and Sharon Ibejih and Shivanya Shomir Dutta and Siddhant Gupta and Silvia Paniagua Suárez and Sina Ahmadi and Sivasuthan Sukumar and Siyuan Song and Snegha A and Sokratis Sofianopoulos and Sona Elza Simon and Sonja Benčina and Sophie Gvasalia and Sphurti More and Spyros Dragazis and Stefan Milosavljević and Stephan P. Kaufhold and Suba S and Sultan Alrashed and Surangika Ranathunga and Taiga Someya and Taja Kuzman Pungeršek and Tal Haklay and Tasi'u Jibril and Tatsuya Aoyama and Tea Abashidze and Terenz Jomar Dela Cruz and Terra Blevins and Themistoklis Nikas and Theresa Idoko and Thu Mai Do and Tilek Chubakov and Tina Munda and Tobiloba Owoeye and Tommaso Gargiani and Uma Rathore and Uni Johannesen and Uwuma Ugwu and Vallerie Alexandra Putra and Vanya Bannihatti Kumar and Varvara Arzt and Vasily Konovalov and Vasudevan Nedumpozhimana and Viktoria Ondrejova and Viktoryia Horbik and Vishnu Vardhan Reddy Kummitha and Vuk Dinić and Walelign Sewunetie and Winston Wu and Xiaojing Zhao and Yacouba Diarra and Yaniv Nikankin and Yash Mathur and Yash Bagla and Yeshil Bangera and Yixi Chen and Yiyuan Li and Yolanda Xavier and Yonatan Belinkov and Zaid Alyafeai and Zhargal Batozargalova and Zhengyang Shan and Zhi Rui Tam and Zilu Tang and Zuzana Nadova and Baber Abbasi and Stella Biderman and David Stap and Duygu Ataman and Fabian Schmidt and Hila Gonen and Jiayi Wang and David Ifeoluwa Adelani},
      year={2026},
      eprint={2510.24081},
      archivePrefix={arXiv},
      primaryClass={cs.CL},
      url={https://arxiv.org/abs/2510.24081}, 
}

@misc{nielsen2026sdusdaisybenchmarkdanish,
      title={{SDUs} {DAISY}: A Benchmark for Danish Culture}, 
      author={Jacob Nielsen and Stine L. Beltoft and Peter Schneider-Kamp and Lukas Galke Poech},
      year={2026},
      eprint={2601.19930},
      archivePrefix={arXiv},
      primaryClass={cs.CL},
      url={https://arxiv.org/abs/2601.19930}, 
}

@misc{smart2026multiwikiqareadingcomprehensionbenchmark,
      title={MultiWikiQA: A Reading Comprehension Benchmark in 300+ Languages}, 
      author={Dan Saattrup Smart},
      year={2026},
      eprint={2509.04111},
      archivePrefix={arXiv},
      primaryClass={cs.CL},
      url={https://arxiv.org/abs/2509.04111}, 
}

@inproceedings{deutsch2025wmt24++,
  title={WMT24++: Expanding the language coverage of WMT24 to 55 languages \& dialects},
  author={Deutsch, Daniel and Briakou, Eleftheria and Caswell, Isaac Rayburn and Finkelstein, Mara and Galor, Rebecca and Juraska, Juraj and Kovacs, Geza and Lui, Alison and Rei, Ricardo and Riesa, Jason and others},
  booktitle={Findings of the Association for Computational Linguistics: ACL 2025},
  pages={12257--12284},
  year={2025}
}

@misc{kinch_nordjylland_news_summarization,
  author       = {Kinch, Oliver},
  title        = {Nordjylland News Summarization},
  howpublished = {Hugging Face},
  year = {2023},
  url          = {https://huggingface.co/datasets/alexandrainst/nordjylland-news-summarization}
}

@misc{larsen_ifeval_da,
  author       = {Larsen, Rasmus and S{\o}rensen, Nathalie Hau and Enevoldsen, Kenneth},
  title        = {{IFEval-da}},
  howpublished = {Hugging Face},
  year = {2025},
  url          = {https://huggingface.co/datasets/danish-foundation-models/ifeval-da}
}

@inproceedings{nielsen-2023-scandeval,
    title = "{S}cand{E}val: A Benchmark for {S}candinavian Natural Language Processing",
    author = "Nielsen, Dan",
    editor = {Alum{\"a}e, Tanel  and
      Fishel, Mark},
    booktitle = "Proceedings of the 24th Nordic Conference on Computational Linguistics (NoDaLiDa)",
    month = may,
    year = "2023",
    address = "T{\'o}rshavn, Faroe Islands",
    publisher = "University of Tartu Library",
    url = "https://aclanthology.org/2023.nodalida-1.20/",
    pages = "185--201"
}

@article{geiping2026scaling,
  title={Scaling up test-time compute with latent reasoning: A recurrent depth approach},
  author={Geiping, Jonas and McLeish, Sean and Jain, Neel and Kirchenbauer, John and Singh, Siddharth and Bartoldson, Brian and Kailkhura, Bhavya and Bhatele, Abhinav and Goldstein, Tom},
  journal={Advances in Neural Information Processing Systems},
  volume={38},
  pages={41340--41391},
  year={2025}
}

@article{kohli2026loop,
  title={Loop, think, \& generalize: Implicit reasoning in recurrent-depth transformers},
  author={Kohli, Harsh and Parthasarathy, Srinivasan and Sun, Huan and Yao, Yuekun},
  journal={arXiv preprint arXiv:2604.07822},
  year={2026}
}

@inproceedings{liu2025toolace,
  title     = {{ToolACE}: Winning the Points of {LLM} Function Calling},
  author    = {Liu, Weiwen and Huang, Xu and Zeng, Xingshan and Hao, Xinlong and Yu, Shuai and others},
  booktitle = {International Conference on Learning Representations (ICLR)},
  year      = {2025},
  note      = {arXiv:2409.00920}
}

@inproceedings{liu2024apigen,
  title     = {{APIGen}: Automated Pipeline for Generating Verifiable and Diverse Function-Calling Datasets},
  author    = {Liu, Zuxin and Hoang, Thai and Zhang, Jianguo and Zhu, Ming and Lan, Tian and others},
  booktitle = {Advances in Neural Information Processing Systems (NeurIPS), Datasets and Benchmarks Track},
  year      = {2024},
  note      = {arXiv:2406.18518}
}

@article{lu2025deepdive,
  title   = {{DeepDive}: Advancing Deep Search Agents with Knowledge Graphs and Multi-Turn {RL}},
  author  = {Lu, Rui and Hou, Zhenyu and Wang, Zihan and Zhang, Hanchen and Liu, Xiao and Li, Yujiang and Feng, Shi and Tang, Jie and Dong, Yuxiao},
  journal = {arXiv preprint arXiv:2509.10446},
  year    = {2025}
}

@article{nvidia2025nemotron3nano,
  title   = {Nemotron 3 Nano: Open, Efficient Mixture-of-Experts Hybrid {Mamba}-{Transformer} Model for Agentic Reasoning},
  author  = {{NVIDIA}},
  journal = {arXiv preprint arXiv:2512.20848},
  year    = {2025}
}

@misc{qwen3.5,
  title = {{Qwen3.5}: Towards Native Multimodal Agents},
  author = {{Qwen Team}},
  month = {February},
  year = {2026},
  url = {https://qwen.ai/blog?id=qwen3.5},
  note = {Accessed: 2026-09-25}
}
\bibliographystyle{iclr2027_conference}
\appendix
\section{Details of the Mimir Training Corpus}
\label{app:mtc}
\begin{table}[h!]
\centering
\caption{Token share by data form.}
\label{tab:form}
\resizebox{\textwidth}{!}{
\begin{tabular}{lrrrrrrr}
\toprule
& Reformatted & Curated + reformatted & Synthetic + audited & Tool-call formatted & Translated + audited & Agreement-supplied & Derived task \\
\midrule
\textbf{Tokens/epoch} & 46.49B & 11.92B & 7.81B & 1.87B & 1.59B & 0.67B & 0.13B \\
\textbf{Share}        & 65.96\% & 16.91\% & 11.08\% & 2.65\% & 2.26\% & 0.95\% & 0.18\% \\
\bottomrule
\end{tabular}
}
\end{table}
\begin{table}[htbp]
\small
\centering
\caption{Token share by language.}
\label{tab:language}
\begin{tabular}{lrrrr}
\toprule
\textbf{Language} & \textbf{English (en)} & \textbf{Danish (da)} & \textbf{Bilingual da+en} & \textbf{Other} \\
\midrule
Tokens/epoch & 48.36B & 17.44B & 4.61B & 0.14B \\
Share        & 68.62\% & 24.74\% & 6.54\% & 0.20\% \\
\bottomrule
\end{tabular}
\end{table}
\subsection{Data Processing and Harmonization}
Datasets enter the corpus in seven distinct forms, reflecting the pipeline's processing stages. We report the forms in Table \ref{tab:form}. \textbf{Reformatted} datasets are existing Hugging Face repositories simply converted into the training format -- this represents the default pathway for public data. \textbf{Curated + reformatted} applies to the Sapient mega-repository, whose 107 sub-collections were selected as a curated subset before reformatting. \textbf{Synthetic + audited} data is LLM-generated using Gemma4 31B and quality-audited before inclusion, with acceptance rates ranging from single digit percentages to high nineties for different categories. This includes transformations from high-quality English and Danish text corpora into span-filling, denoising, reordering, and continuation tasks. Here, we employ Common Pile~\citep{kandpal2025commonpilev018tb} for English text corpora and Danish Dynaword~\citep{6c685239843b47bfb7f47a78aae67929} for Danish ones. We similarly generated and audited the \texttt{mtc1-synthetic-*} instruction datasets and Sapient-synth transplants. \textbf{Tool-call formatted} data incorporates native tool-calling structure for agentic training. \textbf{Translated + audited} covers the OpenHermes-based data and DA/EN translations that were repaired and audited. A small corpus of \textbf{agreement-supplied} data comes from agreements, where licensing does not permit public sharing. Note that we use such datasets to derive synthetic instruction-tuning data rather than train directly on the data. \textbf{Derived task} data is derived from an existing dataset to create a new task formulation.
\begin{table}
\small
\centering
\caption{Top 10 datasets by sampled tokens per epoch.}
\label{tab:top}
\begin{tabular}{lrr}
\toprule
\textbf{Dataset} & \textbf{Tokens/epoch} & \textbf{Share} \\
\midrule
sapientinc/HRM-Text-data-io-cleaned-20260515 & 11.92B & 16.91\% \\
danish-foundation-models/laerebogen         & 8.32B  & 11.81\% \\
nvidia/OpenMathInstruct-2                   & 6.60B  & 9.37\% \\
nvidia/Nemotron-SFT-Agentic-v2              & 4.27B  & 6.06\% \\
danish-foundation-models/dfm-dyna-instruct  & 3.54B  & 5.03\% \\
allenai/Dolci-Instruct-SFT-No-Tools          & 3.49B  & 4.95\% \\
ANONYMIZED/opus-da-en-permissive       & 2.90B  & 4.12\% \\
allenai/Dolci-Instruct-SFT                   & 2.24B  & 3.17\% \\
nvidia/AceReason-1.1-SFT                    & 1.95B  & 2.76\% \\
allenai/big-reasoning-traces                & 1.66B  & 2.35\% \\
\bottomrule
\end{tabular}
\end{table}
\subsection{Details of Task Category Distribution}
The three largest categories: Danish instruction \& knowledge (22.07\%), English instruction (19.26\%), and selected datasets from the Sapient mixed collection (17.02\%)\footnote{\texttt{sapientinc/HRM-Text-data-io-cleaned-20260515}}, together account for over 58\% of the corpus. Math \& reasoning contributes a further 14.8\%, bringing the combined share of the top four categories to 73\%.
The Danish instruction \& knowledge category is the largest by token volume, driven primarily by \texttt{lærebogen}, a Danish instruction-following dataset contributing 8.32B tokens (11.8\% of the corpus) at 4$\times$ repetition, alongside \texttt{dfm-dyna-instruct} (3.54B) and \texttt{synquid\_wiki-instruct-da} (0.99B). English instruction is anchored in Dolci~\citep{olmo2025olmo3} (7.71B combined), Tulu~3~\citep{lambert2024tulu} variants (1.57B), and Nemotron instruction-following data (1.60B). The Sapient mixed category is dominated by a single large repository (11.92B, 16.9\%), which bundles 107 sub-collections from Flan, Platypus, and tasksource. An additional 70 \emph{Sapient-synth} transplant datasets contribute 75M tokens. These are synthetic recreations of English instruction tasks (Flan NIV2, Flan Dialog, Platypus, Tasksource) in generated-and-audited form, replacing original Sapient data that was non-compliant with our permissible data philosophy. Math \& reasoning is led by \texttt{OpenMathInstruct-2}~\citep{toshniwal2024openmath2} (6.60B, 9.4\%), the third-largest single dataset overall, followed by AceReason-1.1-SFT~\citep{liu2025acereason} (1.95B) and verifiable reasoning traces (0.68B). The remaining four categories -- synthetic, agentic \& tool use, machine translation, and science \& summarization -- together contribute 19.0B tokens (27\%).
\subsection{Data Concentration and Repetition}
Table~\ref{tab:top} shows the top 10 datasets by sampled tokens per epoch. The corpus is highly concentrated: these ten datasets account for 66.5\% of all tokens, and the top three alone for 38.1\%, while the remaining 151 datasets contribute 33.5\%. Two single sources exceed 10\% each: the Sapient mega-repository (16.9\%) and lærebogen (11.8\%). The Dolci family contributes 5.73B tokens across two instruction datasets, the largest English instruction contribution after the cleaned Sapient corpus, and NVIDIA sources account for 12.82B tokens across the reasoning, math, and agentic datasets in the top 10.
This concentration is partly by construction: Several datasets are sampled more than once per epoch. Lærebogen is repeated $4\times$, inflating 2.08B base tokens to 8.32B and making it the second-largest entry in the corpus. Eight small Danish datasets are repeated 10x to ensure sufficient coverage despite limited source data, and Dolci-Instruct-SFT-No-Tools is doubled to increase its English instruction contribution. The most heavily repeated dataset, kaenguruen ($20\times$), is negligible in volume at 638K tokens.
\section{Hyperparameters}\label{app:hparams}
\begin{table}[htbp]
\centering
\caption{Model hyperparameters.}
\label{tab:model_hparams}
\resizebox{\textwidth}{!}{
\begin{tabular}{lccccccccccccc}
\toprule
Hidden size & Layers & Half layers & Attn.\ heads & Expansion & H cycles & L cycles & BP max steps & BP warmup ratio & Pos.\ emb. & RoPE $\theta$ & Norm & Norm $\epsilon$ \\
\midrule
1{,}536 & 32 & true & 12 & 4 & 2 & 3 & 5 & 0.2 & RoPE & 10{,}000 & pre-norm & $10^{-6}$ \\
\bottomrule
\end{tabular}
}
\end{table}
\subsection{Model Hyperparameters}\label{app:hparams:model}
Table~\ref{tab:model_hparams} lists the hyperparameters of the model.
\begin{table}[h]
\centering
\caption{Training hyperparameters.}
\label{tab:train_hparams}
\resizebox{\textwidth}{!}{
\begin{tabular}{lccccccccc}
\toprule
Base LR & LR warmup & LR min ratio & Adam $\beta_1$/$\beta_2$ & Weight decay & EMA decay & Global batch & Grad.\ accum.\ steps & Seed \\
\midrule
$3 \times 10^{-4}$ & 2{,}000 & 1.0 & 0.9 / 0.95 & 0.1 & 0.9999 & 262{,}144 & 2 & 0 \\
\bottomrule
\end{tabular}
}
\end{table}
\subsection{Training Hyperparameters}\label{app:hparams:train}
Table~\ref{tab:train_hparams} lists the hyperparameters of Mimir v1 training. Continued training for Mimir v1.5 kept all parameters unchanged but added learning rate scaling (as described in the main part), and finished with a cool-down epoch to $10 \times 10^{-5}$.
\section{Memorisation Audit Details}
\label{app:memorisation-audit}
We run two audits independently from each other across four data categories: (A)  synthetic instruction-tuning data generated from seeds originating from sources covered by agreements; (B) instruction-tuning data from Hugging Face sources for which we could not reliably determine for every row whether an applicable opt-out had been exercised; (C) instruction-tuning data derived from Huggingface sources for which we have high confidence that no applicable opt-out had been exercised for any of the rows; and (D) other synthetic and reasoning data for which we identified no suspicion of material copyright-related concern but where no coherent open or licensing or public domain status could be ascertained.
\subsection*{First Audit}
\begin{figure}[h!]
\centering
\includegraphics[width=0.95\textwidth]{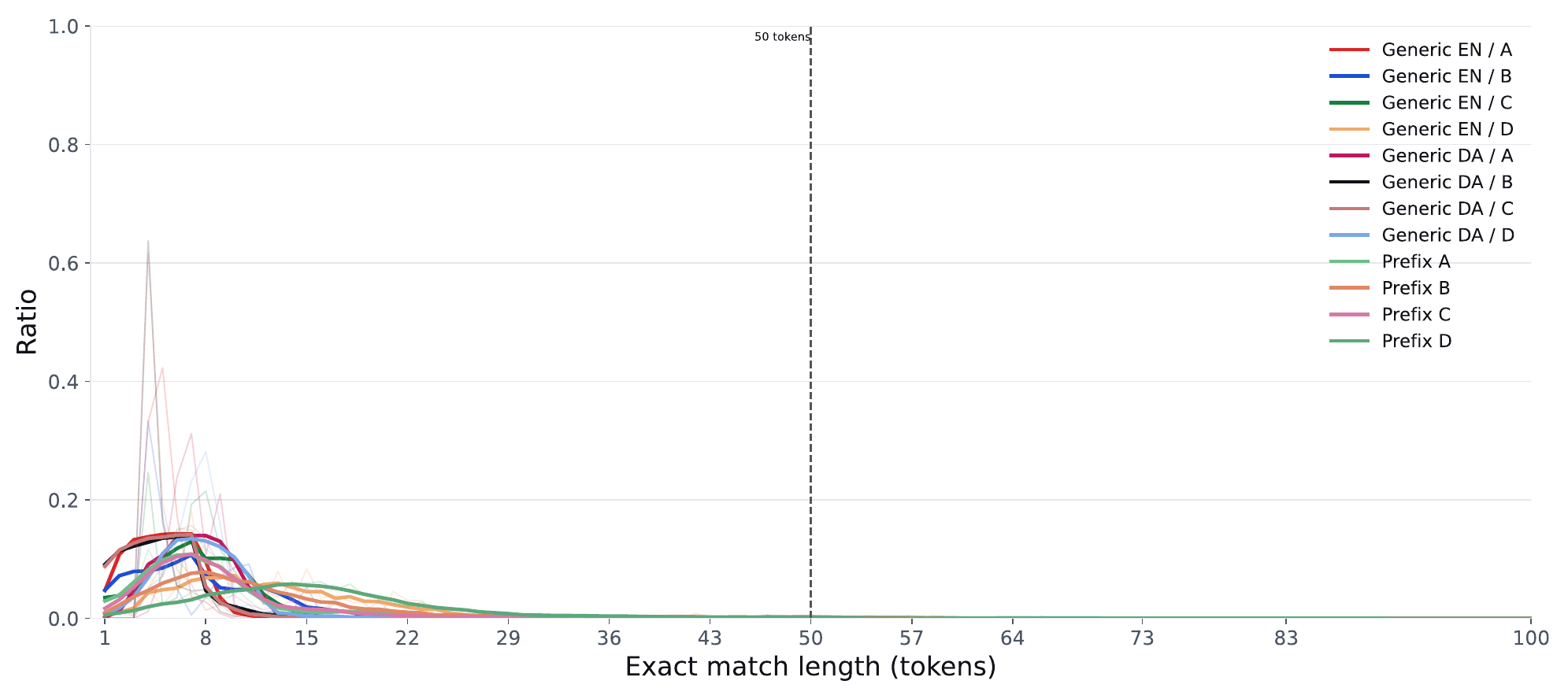}
\caption{Distribution of verbatim matches for each adversarial (A/B/C/D) and non-adversarial (Generic EN, DA) setting. Each non-adversarial setting is applied separately to every dataset. Solid lines are moving averages, faded lines are the raw data.}
\label{fig:exact-match-distribution}
\end{figure}

The first follows the memorisation and propensity methodology of \citet{propme}, evaluating 1{,}000 ordinary non-adversarial prompts (called \emph{generic}) across Danish and English, together with 500 targeted extraction prompts per source category, formed using 50-token source-specific prefixes, for a total of 3{,}000 prompts. The analysis covers 140{,}990{,}504 training documents overall. For each generated output, the method checks against every possible suffix of every document in the analysed collection. Figure~\ref{fig:exact-match-distribution} shows the exact-match length distribution for all settings.
Across the generic settings, the verbatim matches reaching 50 or more tokens are rare: they occur only once in Generic EN/B (0.000056\%) and once in Generic EN/D (0.0000029\%), with none observed in the other English comparisons or any Danish comparison.
Under prefix attacks, verbatim text of any length is identified in 0.073\% of A documents (8{,}731/11{,}989{,}549), 0.887\% of B (15{,}921/1{,}794{,}829), 0.029\% of C (26{,}642/92{,}415{,}358), and 0.046\% of D (16{,}072/34{,}790{,}768).
Detected matches are predominantly short: verbatim spans of 50 tokens or more were found in only 26 (0.00022\%), 271 (0.015\%), 293 (0.00031\%), and 257 (0.00074\%) of all checked training documents across A--D, respectively, and average longest spans are 8.8, 17.1, 12.4, and 22.3 tokens.
Inspection of the matched material shows that A and C include some short natural-language content, while many of the longer matches in B and D consist of numerical sequences, formulas, matrices, code, algorithms, structured tasks, and mathematical problem solving.
\subsection*{Second Audit}
The second audit independently focuses on the type and copyright relevance of reproduced content. Under a fixed greedy 64{+}64-token protocol, it identified 5{,}562 exact-match occurrences across 136{,}612{,}444 model-input evaluations (0.0041\%), corresponding to 3{,}423 unique source-prefix--continuation pairs. By category, this comprised 15/2{,}732{,}080 matches in A (0.0005\%), 7/411{,}508 in B (0.0017\%), 4{,}874/124{,}246{,}748 in C (0.0039\%), and 666/9{,}222{,}108 in D (0.0072\%). These were adjudicated by an LLM and subsequently reviewed by a human for coherence, expressiveness, repetition, and potential copyright significance. Only 61/5{,}562 (1.10\%) contained coherent prose and one (0.018\%) was classified as expressive prose. Thus, the total potentially risky generated outputs with respect to all model-input evaluations are 62/136{,}612{,}444 (0.000045\%).
No high-priority copyright findings were identified, and the single medium-priority case was a predictable continuation of a traditional repetitive song. The remaining prose-like matches were primarily factual captions, news sentences, or text repeated locally within the sources.
\subsection*{Results}
Taken together, these analyses show no evidence of systematic reconstruction of long, distinctive expressive passages and indicate a generally low copyright-related memorisation risk. These results can be assumed to be indicative of a genuine low level of memorisation. They do, however, not fully exclude that extraction under adaptive prompts, longer prefixes, or alternative attack or decoding strategies might be more successful.
\begin{table}[h]
\centering
\caption{Summary of the first memorisation audit. For generic settings, matched documents are reported separately for each language--dataset comparison. The match rate is the share of training documents with any detected verbatim match; the final rate reports the share of all checked training documents containing a verbatim span of at least 50 tokens.}
\label{tab:memorisation_audit_1}
\resizebox{\textwidth}{!}{
\begin{tabular}{lrrrrrr}
\toprule
Setting & Training documents & Matched documents & Match rate & Matches $\geq$50 tokens & $\geq$50-token rate & Avg.\ longest span \\
\midrule
Generic EN / A & 11{,}989{,}549 & 253   & 0.0021\% & 0 & 0.0\% & -- \\
Generic EN / B & 1{,}794{,}829  & 339   & 0.0189\% & 1 & 0.000056\% & -- \\
Generic EN / C & 92{,}415{,}358 & 577   & 0.0006\% & 0 & 0.0\% & -- \\
Generic EN / D & 34{,}790{,}768 & 464   & 0.0013\% & 1 & 0.0000029\% & -- \\
Generic DA / A & 11{,}989{,}549 & 1{,}977 & 0.0165\% & 0 & 0.0\% & -- \\
Generic DA / B & 1{,}794{,}829  & 1{,}710 & 0.0953\% & 0 & 0.0\% & -- \\
Generic DA / C & 92{,}415{,}358 & 2{,}322 & 0.0025\% & 0 & 0.0\% & -- \\
Generic DA / D & 34{,}790{,}768 & 1{,}993 & 0.0057\% & 0 & 0.0\% & -- \\
\midrule
Prefix A & 11{,}989{,}549 & 8{,}731  & 0.073\% & 26  & 0.00022\% & 8.8 tokens \\
Prefix B & 1{,}794{,}829  & 15{,}921 & 0.887\% & 271 & 0.015\% & 17.1 tokens \\
Prefix C & 92{,}415{,}358 & 26{,}642 & 0.029\% & 293 & 0.00032\% & 12.4 tokens \\
Prefix D & 34{,}790{,}768 & 16{,}072 & 0.046\% & 257 & 0.00074\% & 22.3 tokens \\
\bottomrule
\end{tabular}
}
\end{table}
\begin{table}[h]
\centering
\caption{Summary of the second memorisation audit based on exact 64-token continuations and subsequent content-level adjudication. The 5{,}562 exact-match occurrences correspond to 3{,}423 unique source-prefix--continuation pairs.}
\label{tab:memorisation_audit_2}
\resizebox{\textwidth}{!}{
\begin{tabular}{lrrrrr}
\toprule
Category & Model-input evaluations & Exact 64-token matches & Match rate & Coherent prose & Expressive prose \\
\midrule
A & 2{,}732{,}080   & 15      & 0.0005\% & 0  & 0 \\
B & 411{,}508       & 7       & 0.0017\% & 0  & 0 \\
C & 124{,}246{,}748 & 4{,}874 & 0.0039\% & 40 & 0 \\
D & 9{,}222{,}108   & 666     & 0.0072\% & 21 & 1 \\
\midrule
\textbf{Total} & \textbf{136{,}612{,}444} & \textbf{5{,}562} & \textbf{0.0041\%} & \textbf{61 (1.10\%)} & \textbf{1 (0.018\%)} \\
\bottomrule
\end{tabular}
}
\end{table}
\begin{figure}[htbp]
    \centering
    \includegraphics[width=0.8\linewidth]{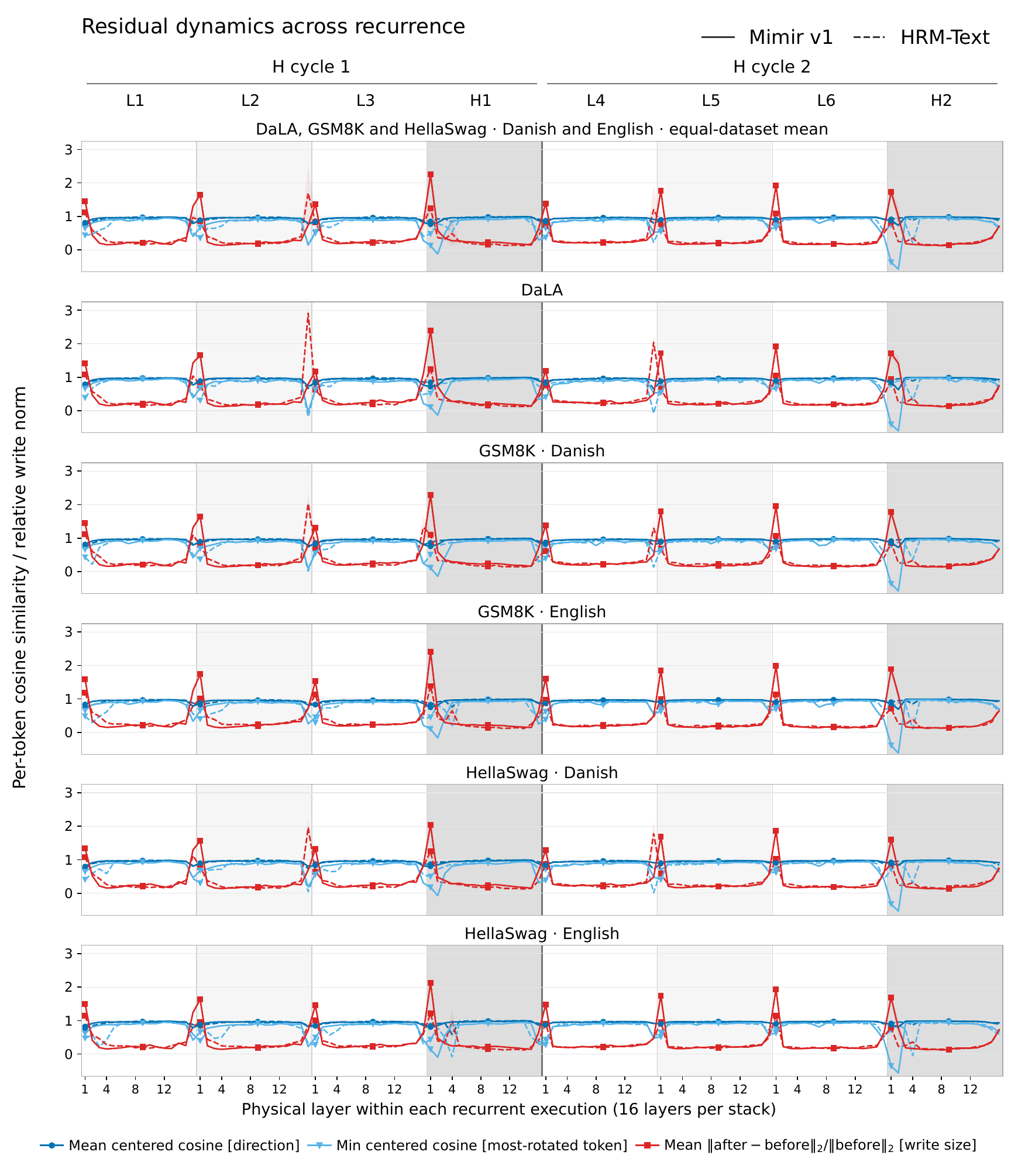}
    \caption{Residual-stream dynamics of Mimir v1 and HRM-Text across DaLA, GSM8K-Danish, GSM8K-English, HellaSwag-Danish and HellaSwag-English. The top panel shows the equally weighted dataset mean; subsequent panels show individual datasets. Measurements include prompt tokens and up to 32 greedily generated continuation tokens. At each physical-layer execution, we measure direction preservation (mean centered cosine similarity), the most-rotated token (minimum centered cosine similarity per sequence, averaged across sequences), and relative write size (the norm of the update divided by the pre-update norm). Both models exhibit recurring peaks in write size and rotation near module boundaries, with broadly consistent patterns across datasets but differences in peak magnitude; lower cosine similarity indicates greater rotation.}
    \label{app:fig:resid}
\end{figure}
\section{Residual Stream Dynamics}\label{app:resid} Figure~\ref{app:fig:resid} shows the residual stream dynamics per dataset measured via relative update magnitude of the residual stream and rotation via centered cosine similarity.
\section{List of Training Datasets for Mimir v1}\label{app:list-of-datasets}
Table~\ref{tab:all_datasets} lists all the datasets with their corresponding HuggingFace identifier, the form type, how many tokens they represent, and the share in percentages.
\begin{longtable}{r p{0.42\textwidth} p{0.18\textwidth} r l}
\caption{All 161 datasets in the Mimir corpus, sorted by sampled tokens per epoch.}\label{tab:all_datasets} \\
\toprule
\textbf{\#} & \textbf{Dataset} & \textbf{Form} & \textbf{Tokens/epoch} & \textbf{Share} \\
\midrule
\endfirsthead
\multicolumn{5}{l}{\textit{continued from previous page}} \\
\toprule
\textbf{\#} & \textbf{Dataset} & \textbf{Form} & \textbf{Tokens/epoch} & \textbf{Share} \\
\midrule
\endhead
1. & sapientinc/HRM-Text-data-io-cleaned-20260515 & Curated + reformatted & 11.92B & 16.91\% \\
2. & danish-foundation-models/laerebogen & Reformatted & 8.32B & 11.81\% \\
3. & nvidia/OpenMathInstruct-2 & Reformatted & 6.60B & 9.37\% \\
4. & nvidia/Nemotron-SFT-Agentic-v2 & Reformatted & 4.27B & 6.06\% \\
5. & danish-foundation-models/dfm-dyna-instruct & Reformatted & 3.54B & 5.03\% \\
6. & allenai/Dolci-Instruct-SFT-No-Tools & Reformatted & 3.49B & 4.95\% \\
7. & schneiderkamplab/opus-da-en-permissive & Reformatted & 2.90B & 4.12\% \\
8. & allenai/Dolci-Instruct-SFT & Reformatted & 2.24B & 3.17\% \\
9. & nvidia/AceReason-1.1-SFT & Reformatted & 1.95B & 2.76\% \\
10. & allenai/big-reasoning-traces & Reformatted & 1.66B & 2.35\% \\
11. & allenai/Dolci-Instruct-SFT-Tool-Use & Tool-call formatted & 1.61B & 2.29\% \\
12. & nvidia/Nemotron-SFT-Instruction-Following-Chat-v2 & Reformatted & 1.60B & 2.27\% \\
13. & allenai/tulu-3-sft-mixture & Reformatted & 1.57B & 2.23\% \\
14. & laion/Scientific-Summaries & Reformatted & 1.27B & 1.80\% \\
15. & schneiderkamplab/common-pile-prefix-continuation & Synthetic + audited & 1.15B & 1.64\% \\
16. & synquid/wiki-instruct-da & Reformatted & 988M & 1.40\% \\
17. & schneiderkamplab/mtc1-openhermes-da & Translated + audited & 922M & 1.31\% \\
18. & schneiderkamplab/common-pile-denoising & Synthetic + audited & 884M & 1.25\% \\
19. & allenai/tulu-v2-sft-mixture & Reformatted & 840M & 1.19\% \\
20. & schneiderkamplab/common-pile-span-filling & Synthetic + audited & 822M & 1.17\% \\
21. & schneiderkamplab/mtc1-openhermes-en & Translated + audited & 672M & 0.95366\% \\
22. & allenai/tulu-v2-sft-long-mixture & Reformatted & 605M & 0.85902\% \\
23. & allenai/verifiable-reasoning-filtered-gpt-41 & Reformatted & 605M & 0.85798\% \\
24. & open-thoughts/OpenThoughts2-1M & Reformatted & 534M & 0.75834\% \\
25. & schneiderkamplab/mtc1-synthetic-native-tool-calling & Synthetic + audited & 513M & 0.728\% \\
26. & schneiderkamplab/transformations-danish-danish & Synthetic + audited & 496M & 0.70403\% \\
27. & schneiderkamplab/transformations-english-english & Synthetic + audited & 471M & 0.66842\% \\
28. & schneiderkamplab/transformations-danish-english & Synthetic + audited & 469M & 0.66529\% \\
29. & schneiderkamplab/transformations-english-danish & Synthetic + audited & 418M & 0.59274\% \\
30. & nvidia/Nemotron-SFT-Multilingual-v1 & Reformatted & 413M & 0.58565\% \\
31. & schneiderkamplab/mtc1-synthetic-danish-summarization-rewrite-controls & Synthetic + audited & 396M & 0.56181\% \\
32. & MegaScience/TextbookReasoning & Reformatted & 374M & 0.5311\% \\
33. & schneiderkamplab/mtc1-synthetic-multiturn-danish-english-chat & Synthetic + audited & 366M & 0.51997\% \\
34. & DBC (agreement-supplied) & Agreement-supplied & 356M & 0.50529\% \\
35. & schneiderkamplab/danish-dynaword-denoising & Synthetic + audited & 323M & 0.45861\% \\
36. & Lex.dk articles & Agreement-supplied & 313M & 0.44467\% \\
37. & oliverkinch/machine-translation-da-en & Reformatted & 284M & 0.40348\% \\
38. & schneiderkamplab/danish-dynaword-prefix-continuation & Synthetic + audited & 252M & 0.35728\% \\
39. & schneiderkamplab/danish-dynaword-span-filling & Synthetic + audited & 251M & 0.35593\% \\
40. & schneiderkamplab/mtc1-synthetic-code-debugging & Synthetic + audited & 247M & 0.34996\% \\
41. & GEM/wiki\_cat\_sum & Reformatted & 205M & 0.29117\% \\
42. & allenai/tulu-3-sft-personas-math & Reformatted & 204M & 0.28922\% \\
43. & giannor/gec\_dala\_tv2r\_it & Reformatted & 193M & 0.27377\% \\
44. & synquid/wildchat-100k-qwen-messages & Reformatted & 190M & 0.2694\% \\
45. & schneiderkamplab/mtc1-synthetic-strict-math-answer-contract & Synthetic + audited & 173M & 0.24605\% \\
46. & schneiderkamplab/mtc1-synthetic-constrained-format-following & Synthetic + audited & 171M & 0.24275\% \\
47. & oliverkinch/danish-summarization & Reformatted & 168M & 0.23888\% \\
48. & schneiderkamplab/common-pile-paragraph-reordering & Synthetic + audited & 167M & 0.23653\% \\
49. & schneiderkamplab/danish-dynaword-paragraph-reordering & Synthetic + audited & 160M & 0.22769\% \\
50. & glaiveai/glaive-function-calling-v2 & Tool-call formatted & 156M & 0.22066\% \\
51. & oliverkinch/machine-translation-da-ar & Reformatted & 140M & 0.19802\% \\
52. & allenai/SciRIFF-train-mix & Reformatted & 132M & 0.18771\% \\
53. & common-pile/arxiv\_papers\_filtered & Derived task & 130M & 0.18386\% \\
54. & oliverkinch/da-instruct-dynaword & Reformatted & 103M & 0.14592\% \\
55. & synquid/translation-100k & Reformatted & 98.2M & 0.13938\% \\
56. & kobprof/skolegpt-instruct & Reformatted & 86.4M & 0.12259\% \\
57. & allenai/verifiable-reasoning-filtered-o4-mini & Reformatted & 80.2M & 0.11378\% \\
58. & oliverkinch/tidsskrift-dk-bt & Reformatted & 76.7M & 0.10878\% \\
59. & allenai/open\_math\_2\_50k\_r1-original & Reformatted & 71.8M & 0.1019\% \\
60. & oliverkinch/machine-translation-da-uk & Reformatted & 69.5M & 0.09859\% \\
61. & giannor/dala\_tv2r\_it & Reformatted & 68.5M & 0.09723\% \\
62. & Salesforce/xlam-function-calling-60k & Tool-call formatted & 67.0M & 0.09502\% \\
63. & oliverkinch/danish-qa & Reformatted & 57.5M & 0.08153\% \\
64. & oliverkinch/dst-table-prompts-bt & Reformatted & 51.1M & 0.07248\% \\
65. & danish-foundation-models/ai\_arena\_udtraek & Reformatted & 45.7M & 0.06483\% \\
66. & oliverkinch/multi-wiki-qa-high-quality-subset & Reformatted & 41.9M & 0.05947\% \\
67. & oliverkinch/dynaword-bt & Reformatted & 34.3M & 0.04867\% \\
68. & HuggingFaceH4/no\_robots & Reformatted & 31.4M & 0.04459\% \\
69. & oliverkinch/da-instruct-dynaword-contemporary-hq & Reformatted & 25.6M & 0.03626\% \\
70. & Team-ACE/ToolACE & Tool-call formatted & 25.2M & 0.03579\% \\
71. & oliverkinch/da-instruct-dynaword-hq & Reformatted & 24.1M & 0.03414\% \\
72. & oliverkinch/danish-university-portals-bt & Reformatted & 21.8M & 0.03088\% \\
73. & allenai/tulu-3-sft-personas-algebra & Reformatted & 21.0M & 0.02982\% \\
74. & allenai/IF\_sft\_data\_verified & Reformatted & 19.9M & 0.02829\% \\
75. & oliverkinch/autodata-da-sft & Reformatted & 18.9M & 0.02677\% \\
76. & oliverkinch/danmarks-statistik-bt & Reformatted & 18.9M & 0.02676\% \\
77. & synquid/danish-verifiable-reasoning & Reformatted & 18.1M & 0.02573\% \\
78. & oliverkinch/da-instruct-dynaword-contemporary & Reformatted & 17.6M & 0.02491\% \\
79. & oliverkinch/eur-lex-bt & Reformatted & 16.7M & 0.02369\% \\
80. & allenai/tulu-3-sft-personas-code & Reformatted & 14.6M & 0.02077\% \\
81. & oliverkinch/instruct-bt & Reformatted & 13.5M & 0.0191\% \\
82. & schneiderkamplab/sapient-synth-flan-dialog-fsopt-data-qrecc & Synthetic + audited & 13.1M & 0.0186\% \\
83. & synquid/ifbench-train & Reformatted & 12.7M & 0.018\% \\
84. & allenai/tulu-3-sft-personas-instruction-following & Reformatted & 11.5M & 0.01634\% \\
85. & schneiderkamplab/sapient-synth-flan-flan-fsopt-data-aeslc-1.0.0 & Synthetic + audited & 10.2M & 0.01441\% \\
86. & synquid/mt-da-deepseek & Reformatted & 8.9M & 0.01261\% \\
87. & allenai/RLVR-MATH & Reformatted & 8.1M & 0.01145\% \\
88. & allenai/RLVR-GSM & Reformatted & 6.8M & 0.00968\% \\
89. & allenai/Dolci-Instruct-SFT-Tool-Use-SA & Tool-call formatted & 5.8M & 0.00825\% \\
90. & schneiderkamplab/sapient-synth-flan-flan-fsnoopt-data-aeslc-1.0.0 & Synthetic + audited & 4.8M & 0.0068\% \\
91. & ccdv/govreport-summarization & Reformatted & 4.4M & 0.00626\% \\
92. & schneiderkamplab/sapient-synth-flan-flan-fsopt-data-opinion-abstracts-rotten-tomatoes & Synthetic + audited & 4.3M & 0.00612\% \\
93. & schneiderkamplab/sapient-synth-flan-dialog-fsopt-data-qrecc-ii & Synthetic + audited & 2.9M & 0.00413\% \\
94. & schneiderkamplab/sapient-synth-flan-niv2-fsopt-data-task589-amazonfood-summary-text-generation & Synthetic + audited & 2.8M & 0.00399\% \\
95. & oliverkinch/eur-lex-sum-instruct & Reformatted & 2.6M & 0.00373\% \\
96. & schneiderkamplab/sapient-synth-flan-niv2-fsopt-data-task590-amazonfood-summary-correction-classification & Synthetic + audited & 2.5M & 0.0036\% \\
97. & schneiderkamplab/sapient-synth-flan-niv2-fsopt-data-task618-amazonreview-summary-text-generation & Synthetic + audited & 2.2M & 0.00309\% \\
98. & schneiderkamplab/sapient-synth-flan-niv2-fsopt-data-task1309-amazonreview-summary-classification & Synthetic + audited & 1.9M & 0.00276\% \\
99. & schneiderkamplab/sapient-synth-flan-flan-zsnoopt-data-aeslc-1.0.0 & Synthetic + audited & 1.9M & 0.00271\% \\
100. & schneiderkamplab/sapient-synth-flan-flan-fsnoopt-data-opinion-abstracts-rotten-tomatoes & Synthetic + audited & 1.7M & 0.00247\% \\
101. & schneiderkamplab/sapient-synth-flan-flan-zsopt-data-aeslc-1.0.0 & Synthetic + audited & 1.7M & 0.00245\% \\
102. & schneiderkamplab/sapient-synth-flan-dialog-zsopt-data-qrecc & Synthetic + audited & 1.7M & 0.00245\% \\
103. & schneiderkamplab/sapient-synth-flan-niv2-fsopt-data-task1375-newscomm-translation & Synthetic + audited & 1.2M & 0.00166\% \\
104. & schneiderkamplab/sapient-synth-flan-niv2-fsopt-data-task1373-newscomm-translation & Synthetic + audited & 1.1M & 0.0015\% \\
105. & schneiderkamplab/sapient-synth-flan-niv2-fsopt-data-task1376-newscomm-translation & Synthetic + audited & 974K & 0.00138\% \\
106. & schneiderkamplab/sapient-synth-platypus-reclor & Synthetic + audited & 922K & 0.00131\% \\
107. & schneiderkamplab/sapient-synth-tasksource-reclor & Synthetic + audited & 901K & 0.00128\% \\
108. & schneiderkamplab/sapient-synth-flan-niv2-fsopt-data-task902-deceptive-opinion-spam-classification & Synthetic + audited & 878K & 0.00125\% \\
109. & schneiderkamplab/sapient-synth-flan-niv2-fsopt-data-task634-allegro-reviews-classification & Synthetic + audited & 816K & 0.00116\% \\
110. & schneiderkamplab/sapient-synth-flan-niv2-fsopt-data-task635-allegro-reviews-answer-generation & Synthetic + audited & 816K & 0.00116\% \\
111. & schneiderkamplab/sapient-synth-flan-niv2-zsopt-data-task590-amazonfood-summary-correction-classification & Synthetic + audited & 811K & 0.00115\% \\
112. & schneiderkamplab/sapient-synth-flan-niv2-zsopt-data-task1309-amazonreview-summary-classification & Synthetic + audited & 795K & 0.00113\% \\
113. & schneiderkamplab/sapient-synth-flan-flan-fsopt-data-opinion-abstracts-idebate & Synthetic + audited & 746K & 0.00106\% \\
114. & schneiderkamplab/sapient-synth-flan-niv2-fsopt-data-task1370-newscomm-classification & Synthetic + audited & 721K & 0.00102\% \\
115. & schneiderkamplab/sapient-synth-flan-niv2-zsopt-data-task589-amazonfood-summary-text-generation & Synthetic + audited & 705K & 0.001\% \\
116. & schneiderkamplab/sapient-synth-flan-niv2-fsopt-data-task1371-newscomm-translation & Synthetic + audited & 703K & 0.001\% \\
117. & oliverkinch/doab-da-bt & Reformatted & 694K & 0.00098\% \\
118. & schneiderkamplab/sapient-synth-flan-niv2-fsopt-data-task1377-newscomm-translation & Synthetic + audited & 679K & 0.00096\% \\
119. & danish-foundation-models/kaenguruen & Reformatted & 638K & 0.0009\% \\
120. & schneiderkamplab/sapient-synth-flan-flan-zsopt-data-opinion-abstracts-rotten-tomatoes & Synthetic + audited & 622K & 0.00088\% \\
121. & schneiderkamplab/sapient-synth-flan-flan-zsnoopt-data-opinion-abstracts-rotten-tomatoes & Synthetic + audited & 616K & 0.00087\% \\
122. & schneiderkamplab/sapient-synth-flan-niv2-fsopt-data-task870-msmarco-answer-generation & Synthetic + audited & 599K & 0.00085\% \\
123. & schneiderkamplab/sapient-synth-flan-niv2-fsopt-data-task1374-newscomm-translation & Synthetic + audited & 599K & 0.00085\% \\
124. & schneiderkamplab/sapient-synth-tasksource-pragmeval-sarcasm & Synthetic + audited & 542K & 0.00077\% \\
125. & schneiderkamplab/sapient-synth-flan-niv2-fsopt-data-task903-deceptive-opinion-spam-classification & Synthetic + audited & 535K & 0.00076\% \\
126. & schneiderkamplab/sapient-synth-flan-niv2-zsopt-data-task618-amazonreview-summary-text-generation & Synthetic + audited & 532K & 0.00076\% \\
127. & schneiderkamplab/sapient-synth-flan-flan-fsnoopt-data-opinion-abstracts-idebate & Synthetic + audited & 527K & 0.00075\% \\
128. & schneiderkamplab/sapient-synth-flan-dialog-zsopt-data-qrecc-ii & Synthetic + audited & 513K & 0.00073\% \\
129. & schneiderkamplab/sapient-synth-flan-niv2-fsopt-data-task265-paper-reviews-language-identification & Synthetic + audited & 395K & 0.00056\% \\
130. & schneiderkamplab/sapient-synth-flan-niv2-zsopt-data-task635-allegro-reviews-answer-generation & Synthetic + audited & 320K & 0.00045\% \\
131. & schneiderkamplab/sapient-synth-flan-niv2-fsopt-data-task909-dialogre-prevalent-speakers & Synthetic + audited & 296K & 0.00042\% \\
132. & schneiderkamplab/sapient-synth-flan-niv2-zsopt-data-task1376-newscomm-translation & Synthetic + audited & 274K & 0.00039\% \\
133. & schneiderkamplab/sapient-synth-flan-niv2-zsopt-data-task1373-newscomm-translation & Synthetic + audited & 268K & 0.00038\% \\
134. & schneiderkamplab/sapient-synth-flan-niv2-zsopt-data-task1375-newscomm-translation & Synthetic + audited & 265K & 0.00038\% \\
135. & schneiderkamplab/sapient-synth-flan-niv2-fsopt-data-task266-paper-reviews-reviewer-perspective & Synthetic + audited & 236K & 0.00033\% \\
136. & schneiderkamplab/sapient-synth-flan-niv2-fsopt-data-task906-dialogre-identify-names & Synthetic + audited & 235K & 0.00033\% \\
137. & schneiderkamplab/sapient-synth-flan-flan-zsnoopt-data-opinion-abstracts-idebate & Synthetic + audited & 223K & 0.00032\% \\
138. & schneiderkamplab/sapient-synth-flan-flan-zsopt-data-opinion-abstracts-idebate & Synthetic + audited & 223K & 0.00032\% \\
139. & schneiderkamplab/sapient-synth-flan-niv2-zsopt-data-task634-allegro-reviews-classification & Synthetic + audited & 215K & 0.0003\% \\
140. & schneiderkamplab/sapient-synth-flan-niv2-zsopt-data-task870-msmarco-answer-generation & Synthetic + audited & 212K & 0.0003\% \\
141. & schneiderkamplab/sapient-synth-flan-niv2-zsopt-data-task902-deceptive-opinion-spam-classification & Synthetic + audited & 198K & 0.00028\% \\
142. & schneiderkamplab/sapient-synth-flan-niv2-fsopt-data-task672-amazon-yelp-summarization & Synthetic + audited & 198K & 0.00028\% \\
143. & schneiderkamplab/sapient-synth-flan-niv2-zsopt-data-task1377-newscomm-translation & Synthetic + audited & 182K & 0.00026\% \\
144. & schneiderkamplab/sapient-synth-flan-niv2-zsopt-data-task1374-newscomm-translation & Synthetic + audited & 174K & 0.00025\% \\
145. & schneiderkamplab/sapient-synth-flan-niv2-zsopt-data-task1371-newscomm-translation & Synthetic + audited & 171K & 0.00024\% \\
146. & schneiderkamplab/sapient-synth-flan-niv2-zsopt-data-task265-paper-reviews-language-identification & Synthetic + audited & 154K & 0.00022\% \\
147. & schneiderkamplab/sapient-synth-flan-niv2-zsopt-data-task903-deceptive-opinion-spam-classification & Synthetic + audited & 149K & 0.00021\% \\
148. & schneiderkamplab/sapient-synth-flan-niv2-fsopt-data-task264-paper-reviews-accept-reject & Synthetic + audited & 140K & 0.0002\% \\
149. & schneiderkamplab/sapient-synth-platypus-scibench & Synthetic + audited & 104K & 0.00015\% \\
150. & schneiderkamplab/sapient-synth-flan-niv2-zsopt-data-task672-amazon-yelp-summarization & Synthetic + audited & 85.3K & 0.00012\% \\
151. & schneiderkamplab/sapient-synth-flan-niv2-zsopt-data-task909-dialogre-prevalent-speakers & Synthetic + audited & 82.0K & 0.00012\% \\
152. & schneiderkamplab/sapient-synth-flan-niv2-zsopt-data-task266-paper-reviews-reviewer-perspective & Synthetic + audited & 71.9K & 0.0001\% \\
153. & schneiderkamplab/sapient-synth-flan-niv2-fsopt-data-task871-msmarco-question-generation & Synthetic + audited & 71.1K & 0.0001\% \\
154. & schneiderkamplab/sapient-synth-flan-niv2-fsopt-data-task908-dialogre-identify-familial-relationships & Synthetic + audited & 68.4K & 0.0001\% \\
155. & schneiderkamplab/sapient-synth-flan-niv2-zsopt-data-task906-dialogre-identify-names & Synthetic + audited & 57.5K & 0.00008\% \\
156. & schneiderkamplab/sapient-synth-flan-niv2-zsopt-data-task264-paper-reviews-accept-reject & Synthetic + audited & 49.3K & 0.00007\% \\
157. & schneiderkamplab/sapient-synth-flan-niv2-fsopt-data-task907-dialogre-identify-relationships & Synthetic + audited & 46.0K & 0.00007\% \\
158. & schneiderkamplab/sapient-synth-flan-niv2-zsopt-data-task871-msmarco-question-generation & Synthetic + audited & 35.6K & 0.00005\% \\
159. & schneiderkamplab/sapient-synth-flan-niv2-zsopt-data-task908-dialogre-identify-familial-relationships & Synthetic + audited & 17.3K & 0.00002\% \\
160. & schneiderkamplab/sapient-synth-flan-niv2-zsopt-data-task907-dialogre-identify-relationships & Synthetic + audited & 13.4K & 0.00002\% \\
161. & schneiderkamplab/sapient-synth-flan-niv2-zsopt-data-task1370-newscomm-classification & Synthetic + audited & 1.9K & $<$0.00001\% \\
\bottomrule
\end{longtable}
\section{Evaluation Setup and Datasets}\label{app:evals}
All models used vLLM-served~\citep{kwon2023efficient} endpoints with FlashInfer, with the exception of Mimir, which requires FlashAttention to correctly capture the PrefixLM and Gemma 4 chat template. We ran both vLLM with FlashAttention4~\citep{zadouri2026flashattention} and Hugging Face Transformers, obtaining comparable results up to numerical stability.
For ease of reproduction, we report the results from Hugging Face Transformers~\citep{wolf2019huggingface}.
Gemma 4 was evaluated in both non-thinking and thinking modes. Thinking requires $\sim$500--650 tokens. MCQ tasks use \texttt{max\_tokens=1}. All non-MCQ tasks (or whenever reasoning is enabled) use \texttt{max\_tokens=2048}.
All baseline evaluations were conducted via the Inspect AI Framework of the~\citet{UK_AI_Security_Institute_Inspect_AI_Framework_2024}.
Table~\ref{tab:nshot} lists all considered evaluation datasets with their precise source as Hugging Face identifier, and the number of in-context shots provided to the models during evaluation. We use the few-shot configuration from \citet{wang2026hrm} the English evaluation datasets, while we use zero-shot evaluation for Math \& Code as well as Danish evaluation datasets.
\begin{table}[htbp]
\centering
\caption{Overview of benchmarks and their configuration. All benchmarks are evaluated with \emph{temperature} 0 and a fixed seed.}
\label{tab:nshot}
\small
\begin{tabular}{llrr}
\toprule
\textbf{Task} & \textbf{Dataset (split)} & \textbf{N-shots} & \textbf{Reference} \\
\midrule
\multicolumn{4}{l}{\textbf{\textit{English}}} \\
BoolQ        & google/boolq          & 5  & \citet{clark2019boolq} \\
Winogrande   & allenai/winogrande    & 5  & \citet{winogrande} \\
Hellaswag    & Rowan/hellaswag       & 10 & \citet{zellers-etal-2019-hellaswag} \\
MMLU         & cais/mmlu             & 5  & \citet{hendrycks2021mmlu} \\
ARC-C        & allenai/ai2\_arc      & 25 & \citet{clark2018think} \\
DROP         & EleutherAI/drop       & 3  & \citet{dua2019dropreadingcomprehensionbenchmark} \\
GovReport    & ccdv/govreport        & 0  & \citet{huang2021govreport} \\
\midrule
\multicolumn{4}{l}{\textbf{\textit{Math \& Code}}} \\
GSM8K        & openai/gsm8k                & 0 & \citet{cobbe2021gsm8k} \\
MATH         & EleutherAI/hendrycks\_math  & 0 & \citet{hendrycks2021math} \\
HumanEval    & human\_eval (164 problems)  & 0 & \citet{chen2021humaneval} \\
\midrule
\multicolumn{4}{l}{\textbf{\textit{Danish}}} \\
Angry Tweets      & DDSC/angry-tweets                    & 0 & \citet{pauli2021danlp_angrytweets} \\
DaLA              & giannor/dala                         & 0 & \citet{barmina2026dala} \\
GEC-DaLA          & giannor/dala\_gen\_v3                & 0 & \citet{barmina2026dala} \\
PIQA-da           & local JSON                           & 0 & \citet{chang2026globalpiqaevaluatingcommonsense} \\
Daisy             & schneiderkamplab/SDU-Daisy          & 0 & \citet{nielsen2026sdusdaisybenchmarkdanish} \\
Multi Wiki QA     & oliverkinch/multi-wiki-qa            & 0 & \citet{smart2026multiwikiqareadingcomprehensionbenchmark} \\
WMT24++ EN-DA     & synquid/wmt24pp                      & 0 & \citet{deutsch2025wmt24++} \\
Nordj. News Summ. & alexandrainst/nordjylland-news-summarization  & 0 & \citet{kinch_nordjylland_news_summarization} \\
IFEval-Da         & danish-foundation-models/ifeval-da   & 0 & \citet{larsen_ifeval_da} \\
Hellaswag-da      & EuroEval                             & 0 & \citet{nielsen-2023-scandeval} \\
\bottomrule
\end{tabular}
\end{table}
\end{document}